\documentclass[runningheads]{llncs}

\usepackage{eccv}

\usepackage{eccvabbrv}

\usepackage{graphicx}
\usepackage{booktabs}

\usepackage[accsupp]{axessibility}

\usepackage{hyperref}

\usepackage{multirow}
\usepackage{caption}
\usepackage[symbol]{footmisc}
\usepackage{marvosym}

\newcommand{\oursfull}{Multi-modal Knowledge Preserving Adapter \xspace}
\newcommand{\ours}{MKP-Adapter\xspace}

\usepackage{xr-hyper}

\begin{document}

\title{Multi-modal Knowledge Preserving Adapter for Embedding Backward Compatibility} 

\titlerunning{MKP-Adapter}
\author{Jaeseok Byun\textsuperscript{1,2\,\Letter} \and
Gukyeong Kwon\textsuperscript{2\,\Letter} \and
Han-Kai Hsu\textsuperscript{2} \and
Meher Gitika Karumuri\textsuperscript{2} \and
Zhikang Zhang\textsuperscript{2} \and
Hao Yang\textsuperscript{2} \and
Davide Modolo\textsuperscript{2\,\Letter}
}
\institute{$^1$Seoul National University, South Korea \\
$^2$Amazon AGI, USA \\[2pt]
\textsuperscript{\Letter}\email{wotjr3868@snu.ac.kr, gukyeong@amazon.com, dmodolo@amazon.com}}

\authorrunning{J.~Byun et al.}

\maketitle
\renewcommand{\thefootnote}{}%
\footnotetext{This work was conducted during Jaeseok Byun’s internship at Amazon.}%
\renewcommand{\thefootnote}{\arabic{footnote}}%

\begin{abstract}

Upgrading embedding models typically requires expensive database re-indexing, as new query embeddings are incompatible with existing database embeddings. While Backward Compatible Training (BCT) mitigates this by enforcing compatibility during training, existing approaches often require updating the backbone model. This is impractical because of significant training cost, the risk of performance regression, and limited access to proprietary model weights. We introduce \oursfull\ (\textbf{\ours}), the first adapter-only BCT approach for Multi-Modal Large Language Models (MLLMs) that requires no backbone updates. We identified that the primary challenge in adapter-only BCT is preserving the knowledge of the new embeddings while enforcing backward compatibility. Hence, we propose a multi-level preservation loss that maintains the geometric structure of the embedding spaces throughout BCT. Furthermore, a focal re-weighting strategy is integrated to prioritize learning from challenging samples. Experiments demonstrate that our method achieves strong backward compatibility across diverse multi-modal benchmarks (image, text, visual document, and video retrieval tasks) and model types. Notably, \ours is trained solely on pre-extracted embeddings and requires only negligible additional latency relative to the original backbone forward pass, highlighting its efficiency.

\keywords{Embedding compatibility \and Multi-modal retrieval \and Adapter}
\end{abstract}

\section{Introduction}
\label{sec:intro}

Embedding models have become fundamental components in modern information retrieval systems, enabling semantic search across diverse data modalities, including text, images, audio, and video \cite{gte,clip,align,audio_retrieval,clip4clip}. However, when a new model is trained independently of an old model for performance improvement, the embeddings produced by the two models are not inherently compatible. This incompatibility forces systems to rebuild their entire embedding database, a process known as ``backfilling'', which typically requires heavy computational overhead, especially for large-scale retrieval systems.

To address this challenge, the Backward Compatible Training (BCT) paradigm \cite{BCT_LCE,BCT_unified,BCT_first,BCT_stanford,BCT_2,BCT_XBT} has been introduced. To be specific, these approaches incorporate explicit backward compatibility constraints into the training objective of the new models and allow the new embeddings to be used for querying databases constructed with old embedding models. Because compatibility constraints must be incorporated during the training stage of the new embedding model, existing BCT methods cannot be applied to pre-trained models and thus require retraining from scratch to achieve compatibility. A recent approach, XBT \cite{BCT_XBT}, attempts to avoid such full retraining through a two-phase procedure: it first trains an adapter on text-only data to bridge old and new embeddings and then applies LoRA-tuning \cite{lora} to the new model backbone using contrastive learning. Although XBT reports promising compatibility results on CLIP \cite{clip} under standard cross-modal retrieval settings, XBT still requires backbone updates.

BCT approaches that require backbone updates, via LoRA-tuning or full retraining, suffer from several crucial limitations, particularly in the context of MLLMs \cite{llava,improved_llava,qwen2-vl,lamra,gme,unime,e5v,vlm2vec,vlm2vec2}.
First, high computational cost. Backbone updates require repeated forward and backward passes and become prohibitively expensive for large-scale modern foundation models.
Second, the risk of degrading model capability.
LoRA-tuning the new model may disturb pre-trained knowledge, while training the new model from scratch with compatibility constraints can compromise the original training objectives as observed by \cite{BCT_first}.
Mitigating these risks requires carefully designed BCT strategies and high-quality large-scale training data. Third, limited applicability due to weight access requirements. Backbone updates require direct access to model weights, making such approaches inapplicable to proprietary API based models that only support black box inference.

\begin{figure}[t!]
\centering
\includegraphics[width=\linewidth]{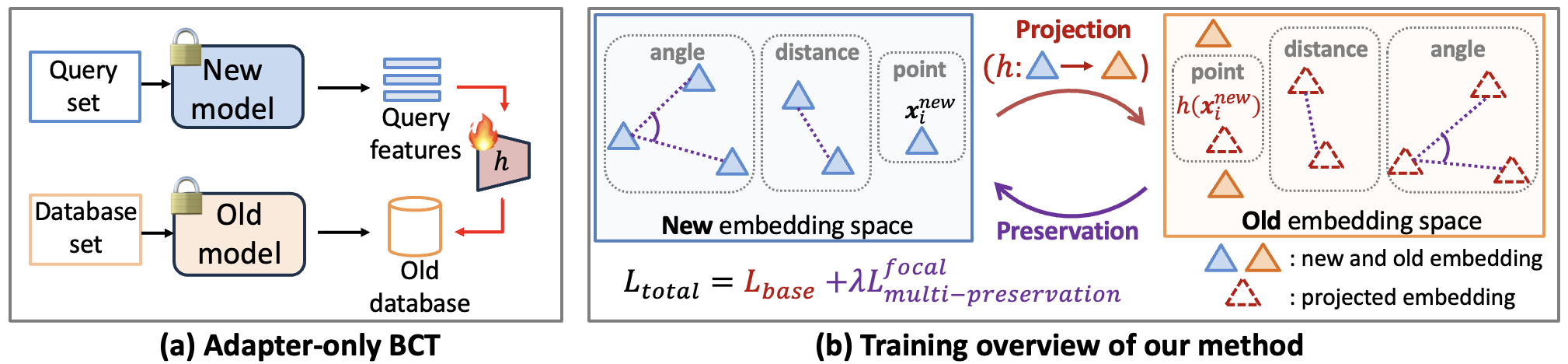} 
\caption{ \textbf{Overview of \ours}. }
\label{fig:overview_ours}
\end{figure}

To that end, we propose a practical alternative that achieves backward compatibility by training only an adapter (shown in \cref{fig:overview_ours}(a)), without requiring any backbone updates. This adapter-only approach resolves aforementioned issues as follows.
First, it operates solely on pre-extracted embeddings and requires no forward or backward passes through the backbone, making it efficient for training and readily applicable to MLLMs.
Second, the pre-trained backbone remains intact, which allows to avoid performance regression and knowledge disturbance.
Third, it requires no access to model weights, enabling direct deployment with API based models.
While this adapter-only approach offers clear practical advantages over prior BCT methods, learning an adapter that achieves strong backward compatibility remains challenging.
Unlike previous BCT approaches that jointly update or LoRA-tune the backbone to enforce compatibility while maintaining discriminative capability, our adapter-only setting requires the adapter alone to satisfy both objectives, with the backbone kept frozen.

A naive approach to obtaining such an adapter is to train it by directly projecting new embeddings onto old ones (\eg, via $L1$ or $L2$ loss). However, forcing projected embeddings to match weaker or noisier old targets can lose the improved representational structure of the new model, which in turn can undermine backward compatibility performance.
Our key idea is therefore to preserve the knowledge encoded in the new embeddings after projection. 
As illustrated in \cref{fig:overview_ours}(b), we achieve this by combining the base projection loss (shown in red) with multi-level preservation regularization (shown in purple) at three granularities: 
(1) \emph{point-wise}, a cycle-consistency loss that reconstructs each original embedding from its projected embedding; 
(2) \emph{distance-wise}, a loss that preserves relative distances between embedding pairs; and 
(3) \emph{angle-wise}, a loss that preserves angular relations among embedding triplets. 
Finally, to account for varying preservation difficulty across modalities and semantic contexts, we apply a focal re-weighting scheme that emphasizes harder samples.

We conduct a comprehensive evaluation under two practical model upgrade scenarios: 
(1) upgrading from dual-tower models (\eg, CLIP\cite{clip}) to MLLMs (\eg, GME\cite{gme}), and (2) upgrading between different versions of MLLMs. 
Our method achieves effective backward compatibility across diverse model upgrade scenarios on standard image-text retrieval benchmarks, including MS-COCO \cite{coco2014}, Flickr30K \cite{flickr30k}, and Urban1K \cite{urban1k}.
We further show that our approach generalizes across diverse tasks and modalities, including instruction-following queries in MMEB \cite{vlm2vec}, visual document retrieval on ViDoRe \cite{vidore}, and video retrieval on MSR-VTT \cite{msrvtt}, VATEX \cite{vatex}, and MSVD \cite{msvd}. 
Across all benchmarks, our method consistently and considerably outperforms baseline approaches in terms of backward compatibility. 
Moreover, the adapter adds only negligible overhead to the total system inference pipeline (0.01\% of the original backbone forward pass cost) and can be trained within 1–2 hours on a moderate-scale (558k paired) dataset (see \cref{sec:analyses}) on 8 A100 GPUs, demonstrating the practicality and efficiency of our approach.
Overall, our contributions are summarized as follows:
\begin{enumerate}
\item We introduce the first adapter-only BCT framework across multiple modalities (image, text, visual documents, and video), particularly suited for using MLLMs as query encoders.
\item We propose a multi-level knowledge preservation loss to maintain the representation quality of the embeddings when training for backward compatibility. Our comprehensive evaluation validates the effectiveness of the proposed loss on diverse modalities and model upgrade scenarios.
\item \ours can be trained within a few hours using pre-extracted embeddings and incurs only negligible overhead to the overall inference pipeline.
\end{enumerate}

\section{Related Work}
\label{sec:related}

\subsection{Multi-modal embedding models} 
Early multi-modal embedding models \cite{clip,align,albef,(GRIT-VLP)byun2022grit,MAFA,siglip} adopt dual-tower architectures trained on large-scale noisy web image-text data. Following LLaVA \cite{llava}, which connects a large language model to a frozen visual backbone, various MLLM variants \cite{qwen2-vl,improved_llava,llavanext,kosmos2,minigpt4} have shown strong performance on multi-modal generation tasks. Several works have since extended MLLMs to embedding-based retrieval: E5-V \cite{e5v} fine-tunes MLLMs with contrastive learning on text-only data, while GME \cite{gme} leverages large-scale image, text, and image-text composition data. VLM2Vec \cite{vlm2vec} is trained on a carefully curated dataset called the Massive Multimodal Embedding Benchmark (MMEB) training set and achieves strong performance on the corresponding MMEB evaluation benchmark.  
UniME \cite{unime} further improves by using knowledge distillation and hard negative mining. 
In this work, we consider compatibility scenarios spanning these diverse MLLM-based query encoders.

\subsection{Embedding compatibility}
Two representative paradigms have been proposed for embedding compatibility: forward compatible training (FCT) and backward compatible training (BCT).
The pioneering FCT work \cite{FCT_side} converts old database embeddings to be compatible with a new model via lightweight adapters, but relies on auxiliary features from a separate model, which are often unavailable in practice.
FastFill \cite{FCT_fastfill} considers online backfilling by estimating the appropriate ordering of the backfilling sample. RBT \cite{BCT_unified} updates both query and database embeddings using a unified adapter. Recently, Embedding Converter \cite{FCT_EC} achieves strong forward compatibility for text embeddings by leveraging relative distance information. 
While FCT only requires training a lightweight adapter, it necessitates converting all database embeddings through the adapter, which can incur substantial computational overhead at massive scale database. Lightweight adapters have also been explored for task and domain adaptation \cite{calibrating,dreditor,retrofitting}, but these methods focus on transforming embeddings within a single model rather than maintaining compatibility across model upgrades. 

In contrast, BCT modifies the training procedure of the new model to ensure compatibility with a fixed gallery of old embeddings, avoiding database updates.
However, incorporating backward compatibility constraints can degrade the discriminative capability of the new model \cite{BCT_first}.
Prior works mitigate this by aligning class centers \cite{BCT_LCE}, expanding embedding dimensionality \cite{BCT_2}, or introducing alignment losses \cite{BCT_stanford}.
These methods require compatibility to be enforced during training, making them unsuitable for frozen pre-trained models and largely limited to unimodal settings.
XBT \cite{BCT_XBT} extends BCT to cross-modal settings via LoRA tuning \cite{lora}, enabling application to frozen models, but it still requires updating the backbone parameters.
In contrast, we propose an adapter-only backward compatibility approach that requires neither backbone modification nor weight access, making it well-suited for using MLLMs as query encoders across diverse retrieval scenarios. We note our method is not tied to any specific model architecture or modality, as it operates directly on pre-extracted embeddings.

\section{Problem Settings}
\label{sec:motivation}

\subsection{Problem formulation} \label{subsec:problem_formulation} 
Our goal is to learn a mapping function from the new embedding space to the old embedding space for backward compatibility, defined as $h: \mathbb{R}^{d^{new}} \rightarrow \mathbb{R}^{d^{old}}$.
Let $f_{\text{old}}$ and $f_{\text{new}}$ denote the frozen pre-trained old and new embedding models, respectively. The corresponding embeddings are given by $\mathbf{x}_i^{\text{old}} = f{_\text{old}}(s_i)$ and $\mathbf{x}_i^{\text{new}} = f_{\text{new}}(s_i)$, where $s_i$ denotes a data sample.
The dataset spans multiple modalities (images, text, visual documents, and videos).
To reflect realistic deployment settings, we assume all embeddings are pre-extracted, requiring no additional forward passes through either embedding model during training.
All embeddings are $L2$-normalized before any adapter operations or indexing.

\subsection{Base adapter}
A naive approach to training such an adapter is to learn a mapping $h: \mathbb{R}^{d^{new}} \rightarrow \mathbb{R}^{d^{old}}$  by minimizing the L2 loss (mean squared error) between the projected embedding and the old embedding, such that $h(\mathbf{x}_{i}^{new}) \approx \mathbf{x}_{i}^{old}$. Here, $d^{new}$ and $d^{old}$ denote the dimensionalities of the new and old embedding models, respectively.
Formally, given $N$ samples, we minimize the distance between embeddings:
\begin{equation}  \label{eq:base_loss}
    \mathcal{L}_{base} = \sum_{i=1}^{N} \| h(\mathbf{x}_{i}^{new})- \mathbf{x}_{i}^{old}\|_2^2.
\end{equation}

While we use mean squared error in this formulation, alternative distance metrics such as mean absolute error may also be employed.
We note that while FCT \cite{FCT_EC,FCT_fastfill} applies such a mapping in the opposite direction (old $\rightarrow$ new), our backward compatibility setting reverses this direction by projecting new embeddings into the old space (new $\rightarrow$ old).
For multi-modal retrieval scenarios such as image–text retrieval, we form the training set by aggregating pre-extracted embeddings from different modalities into a unified dataset. Consequently, each mini-batch contains embeddings from multiple modalities, and the adapter $h$ is trained in a modality-agnostic manner using a single shared mapping across modalities. More details and alternative adapter design choices are provided in the supplementary materials (S.M).

\section{Method}
\label{sec:method}

\subsection{Motivation}
Unlike knowledge distillation \cite{kd_survey,rkd} or forward-compatible training~\cite{FCT_EC,FCT_side,FCT_fastfill}, which use newer or stronger embeddings as the supervision target, an adapter-only BCT is inherently more challenging. Specifically, the base loss $\mathcal{L}_{\text{base}}$ in \cref{eq:base_loss} encourages new embeddings to fit the old embedding space as the target, which typically exhibits a weaker and noisier semantic structure than the new one.
Therefore, using old embeddings as the target naively forces the new embeddings to follow this noisy structure, which can sacrifice the discriminative capacity of the new model and potentially lead to poor backward compatibility performance.
Our key insight is that, rather than simply fitting into the old space, the projection should preserve the semantic structure and knowledge encoded in the new embeddings. Building upon the base loss, we add regularization losses that explicitly preserve the representation quality of the embeddings before and after the projection. In particular, we introduce multi-level preservation losses that maintain the refined similarity structure of the new embeddings at increasing geometric complexity, along with a focal re-weighting strategy that prioritizes challenging samples for improving robustness and generalization.

\subsection{Multi-level preservation loss}
\noindent\textbf{[Point-wise preservation loss]}
We first introduce a point-wise preservation loss inspired by cycle consistency \cite{cycle} to ensure that individual embedding information is preserved after projection. The core idea is that if the projected embedding can be reconstructed back to the original, the essential information of the original embedding is preserved.
To achieve this, we introduce an auxiliary adapter $g : \mathbb{R}^{d^{old}} \rightarrow \mathbb{R}^{d^{new}}$ that maps embeddings from the old space back to the new space.
For each embedding, we project the new embedding to the old space via the main adapter $h : \mathbb{R}^{d^{new}} \rightarrow \mathbb{R}^{d^{old}}$, then reconstruct it back to the new space via $g$. By minimizing the reconstruction error, we enforce that $h$ maintains sufficient semantic information needed to recover the original embedding. Formally, we train two adapters jointly with the following point-level loss:
\begin{equation} \label{eq:point_loss}
\mathcal{L}_{point} = \sum_{i=1}^{N} \|g(h(\mathbf{x}_{i}^{new})) - \mathbf{x}_{i}^{new}\|_2^2.
\end{equation}
$g$ is used only for training and $h$ is the learned adapter used during inference.

\noindent\textbf{[Distance-wise preservation loss]}
While the point-wise loss preserves information at the level of individual embeddings, it does not explicitly enforce the preservation of relationships between embeddings, which are crucial for retrieval tasks where relative similarities determine ranking.
Inspired by \cite{FCT_EC,rkd}, we introduce an additional distance-wise preservation loss that maintains relative distances between embeddings, thereby encouraging the model to preserve the relative similarity structure of the original embeddings in the projected space.  
To capture comprehensive pairwise mutual relationships, we follow \cite{FCT_EC} and consider both global distances computed over randomly sampled pairs within each batch, and local distances computed based on the precomputed $m$-nearest neighbors of each sample selected in the \textit{new} embedding space ($m = 100$). Further details are provided in S.M.  
While this formulation resembles techniques used in knowledge distillation \cite{kd_survey,rkd}, a key distinction is that, instead of learning from an external teacher model that typically provides a stronger supervisory signal, we use the original new embeddings themselves as self-targets to preserve their inherent relational structure during projection.
We define the distance-wise structure preservation loss as:
\begin{equation}  \label{eq:distance_loss}
\mathcal{L}_{distance} = \sum_{(i,j) \in \mathcal{P}} \left| Dist(h(\mathbf{x}_{i}^{new}), h(\mathbf{x}_{j}^{new})) - Dist(\mathbf{x}_{i}^{new}, \mathbf{x}_{j}^{new}) \right|,
\end{equation}
where $\mathcal{P}$ denotes anchor–reference pairs $(i,j)$, with $i$ sampled from the mini-batch and $j$ chosen either randomly within the batch (global) or from the $m$ nearest neighbors of $i$ (local). The distance function is defined as $Dist(\mathbf{a}, \mathbf{b}) = 1 - \cos(\mathbf{a}, \mathbf{b})$, where $\cos(\cdot, \cdot)$ denotes cosine similarity.

\noindent\textbf{[Angle-wise preservation loss]}
To further preserve the relative similarity structure, we incorporate an angle-wise loss inspired by relational knowledge distillation (RKD) \cite{rkd}.
While distance-based losses are defined by pairs of embeddings, angle-wise losses incorporate triplets. 
By leveraging the higher-order structural information provided by these additional embeddings, we hypothesize that angle-wise regularization complements distance preservation in maintaining the overall manifold structure.
As with the previous losses (\cref{eq:point_loss,eq:distance_loss}), our formulation is self-supervised; unlike RKD, which uses a strong teacher as the target, we preserve the angles defined by the new embeddings themselves after projection.
Given a triplet of embeddings, the angle formed at the vertex $\mathbf{x}_j$ is defined as:
\begin{equation}
\psi_A(\mathbf{x}_{i}, \mathbf{x}_{j}, \mathbf{x}_{k}) = \cos \angle \mathbf{x}_{i}\mathbf{x}_{j}\mathbf{x}_{k} = \langle \mathbf{e}^{ij}, \mathbf{e}^{kj} \rangle,
\end{equation}
where $\mathbf{e}^{ij} = \frac{\mathbf{x}_{i} - \mathbf{x}_{j}}{\|\mathbf{x}_{i} - \mathbf{x}_{j}\|_2}$ and $\mathbf{e}^{kj} = \frac{\mathbf{x}_{k} - \mathbf{x}_{j}}{\|\mathbf{x}_{k} - \mathbf{x}_{j}\|_2}$ are normalized directional vectors. This measures the angle at $\mathbf{x}_j$ formed by the triplet $(\mathbf{x}_i, \mathbf{x}_j, \mathbf{x}_k)$.

The angle-wise preservation loss then maintains these angular relationships after projection:

\begin{equation} \label{eq:angle_loss}
\mathcal{L}_{angle} = \sum_{(i,j,k) \in \mathcal{T}} \left| \psi_A(h(\mathbf{x}_{i}^{new}), h(\mathbf{x}_{j}^{new}), h(\mathbf{x}_{k}^{new})) - \psi_A(\mathbf{x}_{i}^{new}, \mathbf{x}_{j}^{new}, \mathbf{x}_{k}^{new}) \right|.
\end{equation}
where $\mathcal{T}$ denotes the set of embedding triplets formed using global (random) and local (nearest-neighbor) sampling.

Together, the multi-level preservation loss function is defined as:
\begin{equation} \label{eq:multi_preserve}
\mathcal{L}_{multi-preservation} = \mathcal{L}_{point}+ \mathcal{L}_{distance} + \mathcal{L}_{angle}
\end{equation}

\subsection{Focal re-weighting strategy} 
For more effective training, we employ a focal re-weighting strategy into the multi-level preservation losses (\cref{eq:multi_preserve}). Preservation difficulty might vary across modalities and semantics, with some samples being easily aligned while others remain challenging. Without re-weighting, easy samples can dominate the training loss, preventing the adapter from adequately learning difficult cases. 
Inspired by focal loss in object detection~\cite{focal}, we re-weight the loss term based on errors between original and reconstructed embeddings to emphasize challenging examples. For example, the focal-weighted point-level loss is defined as:

\begin{equation} \label{eq:point_focal_loss}
\begin{split}
\mathcal{L}_{point}^{focal} &= \frac{1}{N}\sum_{i=1}^{N} w_i \Delta_i, \\
\text{where} \quad & \Delta_i = ||g(h(\mathbf{x}_i^{new})) - \mathbf{x}_i^{new}||_2^2, \quad w_i = \left(\frac{\Delta_i}{\bar{\Delta}}\right)^{\gamma}, \quad \bar{\Delta} = \frac{1}{N}\sum_{i=1}^{N} \Delta_i.
\end{split}
\end{equation}
where $\gamma$ is the focusing parameter that controls the strength of re-weighting, with $\gamma=0$ corresponding to no re-weighting and larger values placing more emphasis on harder samples.
We apply the same focal re-weighting strategy to the distance-wise and angle-wise preservation losses defined in \cref{eq:distance_loss,eq:angle_loss}; formal definitions are provided in S.M.

Finally, we obtain our model, dubbed \ours, by optimizing the following training objective, which combines all proposed loss components:
\begin{equation}
\mathcal{L}_{total} = \mathcal{L}_{base} + \lambda \mathcal{L}_{multi-preservation}^{focal},
\end{equation}
where $\mathcal{L}_{\text{multi-preservation}}^{focal}$ denotes the combined multi-level preservation losses with focal re-weighting, and $\lambda$ controls the trade-off between projection onto the old embedding space and preservation of the new embedding capabilities.

\section{Experimental Results}
\label{sec:experiments}
We evaluate the compatibility performance of \ours across diverse upgrade scenarios and modalities. We first demonstrate its effectiveness on standard image-to-text (I2T) and text-to-image (T2I) cross-modal retrieval tasks. Beyond retrieval, we also assess performance on complex MMEB benchmark \cite{vlm2vec}, which includes a range of embedding-dependent tasks such as classification, visual grounding, and visual question answering. We further extend our evaluation to additional modalities such as visual document and video.
We consider two representative model upgrade scenarios from old to new embedding models:
(1) Dual-tower $\rightarrow$ MLLM, upgrading from traditional dual-tower architectures to MLLMs, where CLIP ViT-L/14~\cite{clip,openclip} is used as the old model, and UniME (7B)~\cite{unime} (or GME (7B)~\cite{gme}) as the new model; and
(2) MLLM $\rightarrow$  MLLM, upgrading between MLLM variants, where E5-V (7B)~\cite{e5v} is used as the old model, and UniME (7B)~\cite{unime} as the new model.

\subsection{Experimental settings}
\noindent{\textbf{[Training setup]}} 
To obtain training embeddings for image and text retrieval, we use the LLaVA-LCS image–text paired dataset~\cite{llava}, which contains approximately 558K image–text pairs. From this dataset, we construct three types of training inputs: (1) images only, (2) texts only, and (3) image+text compositions where paired image–text samples are provided jointly as input to the MLLM. The composition inputs are included to better support instruction-following scenarios in the MMEB benchmark, where queries combine both textual and visual information.
As described in \cref{sec:method}, we aggregate all inputs into a unified training set for adapter training. During training, inputs are randomly sampled to form mini-batches across diverse modalities. Additional details and ablations, including results without composition data, are provided in S.M.
We further experiment with larger datasets, SBU~\cite{sbu} (1M pairs) and CC3M~\cite{cc3m} (3M pairs), to analyze the effect of training data scale on compatibility performance.
For visual document retrieval, we use the ColPali training set~\cite{vidore}, which contains approximately 118K paired samples. Each visual document is treated as a single image during both training and evaluation.
For video retrieval, we use a subset of the LLaVA-Hound video dataset~\cite{llavahound} with approximately 300K paired samples. To obtain video embeddings, we uniformly sample 8 frames from each video, extract embeddings for each frame independently, and then average them to form a single video representation, following prior work \cite{internvideo}.

We use a 2-layer MLP as our adapter architecture with the following structure: LayerNorm-Linear-GELU-Dropout(0.1)-Linear.
The default hidden dimension is set to be equal to the input embedding dimension.
We train the adapter using the AdamW optimizer \cite{adamw} with a weight decay of $10^{-4}$, a batch size of 32, and an initial learning rate of $10^{-3}$.
A cosine annealing learning rate schedule is applied, decaying the learning rate from the initial value to zero, without warmup.
All experiments are conducted on 8 A100 (40GB) GPUs.
We use the $L2$ loss as the default base projection loss defined in \cref{eq:base_loss}, although our preservation regularization can be combined with alternative base losses. 
Both $\gamma$ and $\lambda$ are set to 1.0 by default.
Additional training details are provided in S.M. 
Results under diverse hyperparameter choices are reported in \cref{tables/analyses_batch,tables/ablation_hyperparameter}.

\noindent{\textbf{[Evaluation setup]}} 
We evaluate the backward compatibility performance of \ours across multiple benchmarks, where query embeddings are computed as $h(\mathbf{x}^{new})$ using the adapter trained by each method, while database embeddings remain fixed as $\mathbf{x}^{old}$ from the old model. We follow the official evaluation protocols and metrics of each benchmark for comparability with prior works. First, we demonstrate its effectiveness on standard I2T and T2I retrieval using MS-COCO \cite{coco2014}, Flickr30K \cite{flickr30k}, and Urban1K \cite{urban1k}, reporting Recall@1. Additional Recall@10 results are provided in \cref{tables/i2t_t2i_R_10}, and large-scale evaluation results on the full COCO validation set are reported in \cref{tables/coco_val_whole}.
Second, we conduct a comprehensive evaluation on MMEB, a more challenging benchmark comprising 36 datasets across four task categories (classification, VQA, retrieval, and visual grounding).
We note this evaluation assesses the generalization ability of the adapters across diverse tasks and domains, as MMEB includes sub-datasets drawn from domains substantially different from the training data. 
Moreover, the queries contain explicit instructions, allowing us to evaluate instruction-following capability as well. 
Following standard protocols, we report Precision@1 as the metric. 
Third, we evaluate text-to-visual-document retrieval on ViDoRe (V1) \cite{vidore}, a benchmark of 10 sub-datasets spanning various content types (text, figures, infographics, tables) and domains (medical, business, scientific, administrative). Following prior work \cite{vidore}, we report nDCG@5 as the evaluation metric.
Finally, we assess video retrieval performance on three benchmarks: MSR-VTT \cite{msrvtt}, MSVD \cite{msvd}, and VATEX \cite{vatex}, using Recall@1 as the metric. We report results only for CLIP$\rightarrow$GME on document and video retrieval benchmarks, as UniME is not suitable as the new model due to its weak performance on both benchmarks (see \cref{sec:appendix_exclude}).

\begin{figure}[t!]
\centering
\includegraphics[width=\linewidth]{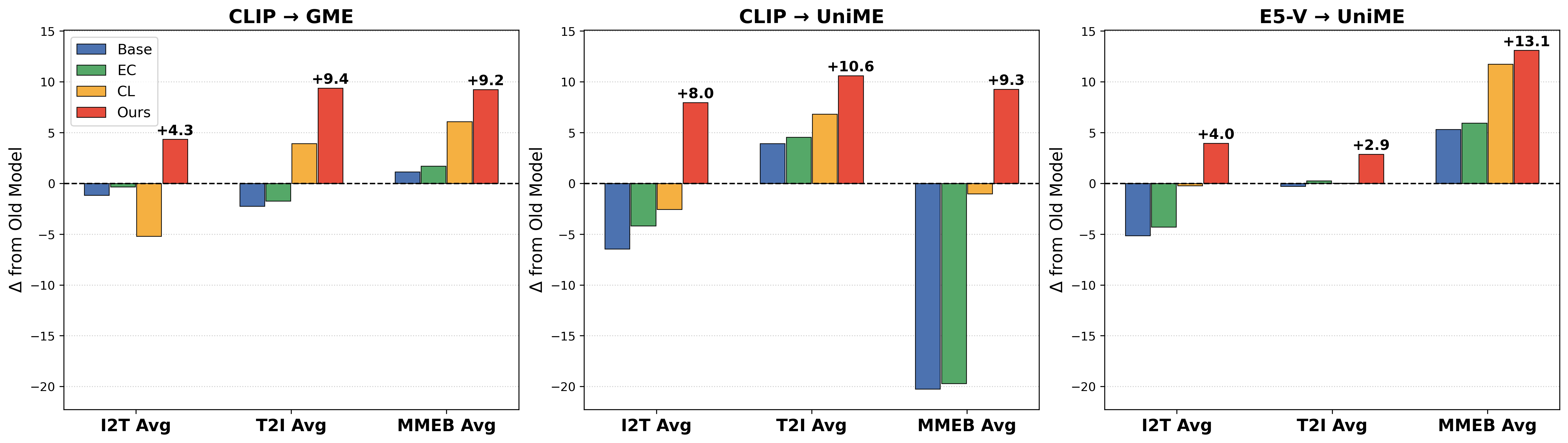} 
\caption{\textbf{Histogram of main results.} The y-axis shows the performance gap relative to the old model; the black dashed line at zero indicates the old model's performance. ``I2T/T2I AVG'' and ``MMEB AVG'' correspond to the average scores computed over the benchmarks listed in \cref{tables/i2t_t2i_table} and \cref{tables/mmeb_table}, respectively.
}
\label{fig:results_histogram}
\end{figure}

\begin{table}[t!]
    \centering
     \caption{\textbf{Compatibility results on standard I2T/T2I benchmarks (MS-COCO \cite{coco2014}, Flickr30K \cite{flickr30k}, Urban1K \cite{urban1k}.)} $\mathbf{x}^{new}$ and $\mathbf{x}^{old}$ denote embeddings from the new and old models, respectively.
    $h$ denotes the adapter function trained by each method. We report Recall@1 as the evaluation metric. \textbf{Bold} indicates the best performance among BCT methods.}
     \resizebox{\columnwidth}{!}{
    \setlength{\tabcolsep}{3pt}
    
\begin{tabular}{ccc|cccccccc}
\midrule
\multicolumn{2}{c|}{\textbf{Case}}         & \multirow{2}{*}{\textbf{Method}} & \multicolumn{2}{c}{\textbf{COCO}} & \multicolumn{2}{c}{\textbf{Flickr30K}} & \multicolumn{2}{c|}{\textbf{Urban1K}} & \multicolumn{2}{c}{\textbf{Average}} \\ \cline{4-11} 
query      & \multicolumn{1}{c|}{database} &                                  & I2T R@1         & T2I R@1         & I2T R@1          & T2I R@1          & I2T R@1 & \multicolumn{1}{c|}{T2I R@1} & I2T R@1                  & T2I R@1   \\ \midrule
\multicolumn{11}{c}{$\mathbf{x}^{old}$: CLIP ViT-L/14 \cite{openclip}, $\mathbf{x}^{new}$: GME-7B \cite{gme}}                                                                                                                                                                                    \\ 
 \midrule
$\mathbf{x}^{old}$     & \multicolumn{1}{c|}{$\mathbf{x}^{old}$}   & -                                & 56.32           & 36.50           & 85.10            & 65.00            & 68.00   & \multicolumn{1}{c|}{55.90}   & 69.80                    & 52.47     \\
$\mathbf{x}^{new}$     & \multicolumn{1}{c|}{$\mathbf{x}^{new}$}   & -                                & 68.68           & 57.35           & 90.80            & 80.90            & 90.10   & \multicolumn{1}{c|}{88.60}   & 83.20                     & 75.61     \\ \midrule
$h(\mathbf{x}^{new})$  & \multicolumn{1}{c|}{$\mathbf{x}^{old}$}   & Base adapter \cite{FCT_side}                     & 58.38           & 34.93           & 85.50             & 64.74            & 62.00   & \multicolumn{1}{c|}{51.00}   & 68.63                   & 50.22     \\
$h(\mathbf{x}^{new})$  & \multicolumn{1}{c|}{$\mathbf{x}^{old}$}   & EC-style adapter \cite{FCT_EC}                 & 59.52          & 35.88          & 85.30            & 65.74            & 63.50   & \multicolumn{1}{c|}{50.60}   & 69.44                    & 50.74     \\
$h(\mathbf{x}^{new})$  & \multicolumn{1}{c|}{$\mathbf{x}^{old}$}   & CL-based adapter \cite{seo2025metric}                     & 54.38               & 37.06               & 76.00                & 68.00               & 63.40      & \multicolumn{1}{c|}{64.10}       & 64.59                        & 56.38      \\
\rowcolor{gray!20} $h(\mathbf{x}^{new})$  & \multicolumn{1}{c|}{$\mathbf{x}^{old}$}   & \ours                         &  \textbf{66.18}          &  \textbf{38.78}          &  \textbf{87.50}           &  \textbf{69.16}            &  \textbf{68.70}   & \multicolumn{1}{c|}{ \textbf{77.60}}   &  \textbf{74.13}                    &  \textbf{61.85}    \\ \midrule

\multicolumn{11}{c}{$\mathbf{x}^{old}$: CLIP ViT-L/14 \cite{openclip}, $\mathbf{x}^{new}$: UniME-7B \cite{unime}}                                                                                                       
\\ \midrule
$\mathbf{x}^{old}$     & \multicolumn{1}{c|}{$\mathbf{x}^{old}$}   & -                                & 56.32           & 36.50           & 85.10            & 65.00            & 68.00   & \multicolumn{1}{c|}{55.90}   & 69.80                    & 52.47    \\
$\mathbf{x}^{new}$     & \multicolumn{1}{c|}{$\mathbf{x}^{new}$}   & -                                 & 70.08           & 53.71           & 93.40            & 82.34            & 96.10   & \multicolumn{1}{c|}{95.40}   & 86.52                    & 77.15    \\ \midrule
$h(\mathbf{x}^{new})$  & \multicolumn{1}{c|}{$\mathbf{x}^{old}$}   & Base adapter \cite{FCT_side}                     & 53.74           & 37.44          & 81.90           & 66.22           & 54.50   & \multicolumn{1}{c|}{65.50}   & 63.35                  & 56.39     \\
$h(\mathbf{x}^{new})$  & \multicolumn{1}{c|}{$\mathbf{x}^{old}$}   & EC-style adapter \cite{FCT_EC}                 & 56.22          & 38.50          & 82.20            & 67.70            & 58.40   & \multicolumn{1}{c|}{64.80}   & 65.61                  & 57.00     \\
$h(\mathbf{x}^{new})$  & \multicolumn{1}{c|}{$\mathbf{x}^{old}$}   & CL-based adapter \cite{seo2025metric}                     & 53.00               & 38.42               & 79.40               & 68.76               & 69.30       & \multicolumn{1}{c|}{70.70}       & 67.23                        & 59.29      \\
\rowcolor{gray!20} $h(\mathbf{x}^{new})$  & \multicolumn{1}{c|}{$\mathbf{x}^{old}$}   & \ours                         & \textbf{66.68}         &  \textbf{42.22}          &  \textbf{91.20}           &  \textbf{72.00}           &  \textbf{75.40}   & \multicolumn{1}{c|}{ \textbf{75.00}}   &  \textbf{77.76 }                   &  \textbf{63.07}     \\ \midrule

\multicolumn{11}{c}{$\mathbf{x}^{old}$: E5-V \cite{e5v}, $\mathbf{x}^{new}$: UniME-7B \cite{unime}}                                                                                                                                                                                        \\ \midrule

$\mathbf{x}^{old}$     & \multicolumn{1}{c|}{$\mathbf{x}^{old}$}   & -                                & 62.24           & 50.98           & 88.20            & 80.64            & 83.80   & \multicolumn{1}{c|}{85.60}   & 78.08                    & 72.41     \\
$\mathbf{x}^{new}$     & \multicolumn{1}{c|}{$\mathbf{x}^{new}$}   & -                                & 70.08           & 53.71           & 93.40            & 82.34            & 96.10   & \multicolumn{1}{c|}{95.40}   & 86.52                    & 77.15     \\ \midrule
$h(\mathbf{x}^{new})$  & \multicolumn{1}{c|}{$\mathbf{x}^{old}$}   & Base adapter \cite{FCT_side}                     & 55.90           & 50.21           & 85.10            & 79.22            & 77.80   & \multicolumn{1}{c|}{86.90}   & 72.93                    & 72.11     \\
$h(\mathbf{x}^{new})$  & \multicolumn{1}{c|}{$\mathbf{x}^{old}$}   & EC-style adapter \cite{FCT_EC}                      & 57.30               & 50.32               & 85.90               & 79.52                & 78.10       & \multicolumn{1}{c|}{88.10}       & 73.77                        & 72.65      \\
$h(\mathbf{x}^{new})$  & \multicolumn{1}{c|}{$\mathbf{x}^{old}$}   & CL-based adapter \cite{seo2025metric}                     & 63.18               & 50.12               & 88.70                & 78.46                & 81.70       & \multicolumn{1}{c|}{88.70}       & 77.86                        & 72.43      \\
\rowcolor{gray!20} $h(\mathbf{x}^{new})$  & \multicolumn{1}{c|}{$\mathbf{x}^{old}$}   & \ours                      & \textbf{65.28}          & \textbf{52.61}           & \textbf{90.60}            & \textbf{81.10}           & \textbf{90.20}   & \multicolumn{1}{c|}{\textbf{92.10}}   & \textbf{82.03}                    & \textbf{75.27}    \\ 

\bottomrule

\end{tabular}
    
    }
    \label{tables/i2t_t2i_table}
\end{table}

\noindent{\textbf{[Baselines]}} To the best of our knowledge, no prior work has explored adapter-only approaches for backward compatibility in multimodal retrieval. We therefore establish three adapter-only baselines for comparison, adapted from related literature, in which only the adapter is trained while both the old and new embedding models remain frozen. 
\begin{itemize}
    \item[(a)] Base adapter \cite{FCT_side}. This adapter is trained using the $L2$ loss defined in \cref{eq:base_loss}. 
We adopt $L2$ instead of $L1$, as it consistently yields superior performance in our experiments (See \cref{tables/analyses_3}).

\item[(b)] EC-style adapter \cite{FCT_EC}. 
Embedding Converter (EC) \cite{FCT_EC} achieves state-of-the-art performance in forward-compatible training (FCT) for text retrieval (old $\rightarrow$ new) by distilling relative distance information between embeddings in addition to the $L2$ loss.
We adopt this method to the reverse direction (new $\rightarrow$ old) to establish a backward-compatibility baseline.

\item[(c)] CL-based base adapter \cite{seo2025metric,BCT_XBT}. This adapter is trained using an InfoNCE-style contrastive objective~\cite{simclr,infonce}:

\begin{table}[t!]
    \centering
     \caption{\textbf{Compatibility results on MMEB benchmarks.} The numbers in parentheses indicate the number of sub-datasets in each category. We report Precision@1 as the evaluation metric. Other details are the same as in \cref{tables/i2t_t2i_table}.}
     \resizebox{\columnwidth}{!}{
    \setlength{\tabcolsep}{3pt}
    \begin{tabular}{ccccccc|c}
\midrule
\multicolumn{2}{c|}{\textbf{Case}}         & \multicolumn{1}{c|}{\multirow{2}{*}{\textbf{Method}}} & \multicolumn{4}{c|}{\textbf{Per Meta-Task Score}}                & \multicolumn{1}{c}{\textbf{Average}}      \\ \cline{4-8} 
query      & \multicolumn{1}{c|}{database} & \multicolumn{1}{c|}{}                                 & Classification (10) & VQA (10) & Retrieval (12) & Grounding (4) & Overall (36) \\ \midrule

\multicolumn{8}{c}{$\mathbf{x}^{old}$: CLIP ViT-L/14 \cite{openclip}, $\mathbf{x}^{new}$: GME-7B \cite{gme}}                                                                                                                                                        \\ \midrule
$\mathbf{x}^{old}$      & \multicolumn{1}{c|}{$\mathbf{x}^{old}$ }   & \multicolumn{1}{c|}{-}                                & 52.80               & 8.88     & 51.45          & 55.52         & 40.29        \\
$\mathbf{x}^{new}$    & \multicolumn{1}{c|}{$\mathbf{x}^{new}$ }   & \multicolumn{1}{c|}{-}                                & 57.12               & 34.87    & 72.10          & 59.95         & 56.25        \\ \midrule
$h(\mathbf{x}^{new})$  & \multicolumn{1}{c|}{$\mathbf{x}^{old}$ }   & \multicolumn{1}{c|}{Base adapter \cite{FCT_side}}                     & 53.99               & 19.28    & 51.18          & 36.15        & 41.43        \\
$h(\mathbf{x}^{new})$  & \multicolumn{1}{c|}{$\mathbf{x}^{old}$ }   & \multicolumn{1}{c|}{EC-style adapter \cite{FCT_EC}}                     & 54.44               & 19.35    & 52.01        & 37.33        & 41.98      \\
$h(\mathbf{x}^{new})$  & \multicolumn{1}{c|}{$\mathbf{x}^{old}$ }   & \multicolumn{1}{c|}{CL-based adapter \cite{seo2025metric}}                & 53.43               & 23.55   & 57.61          & 52.08         & 46.37        \\
\rowcolor{gray!20} $h(\mathbf{x}^{new})$  & \multicolumn{1}{c|}{$\mathbf{x}^{old}$ }   & \multicolumn{1}{c|}{\ours}                      & \textbf{54.88}    & \textbf{27.93}    & \textbf{61.26}          & \textbf{54.85}        & \textbf{49.52}      \\ \midrule

\multicolumn{8}{c}{$\mathbf{x}^{old}$: CLIP ViT-L/14 \cite{openclip}, $\mathbf{x}^{new}$: UniME-7B \cite{unime}}                                                                                                                                                        \\ \midrule
$\mathbf{x}^{old}$      & \multicolumn{1}{c|}{$\mathbf{x}^{old}$ }   & \multicolumn{1}{c|}{-}                                & 52.80               & 8.88     & 51.45          & 55.52         & 40.29        \\
$\mathbf{x}^{new}$    & \multicolumn{1}{c|}{$\mathbf{x}^{new}$ }   & \multicolumn{1}{c|}{-}                                & 60.54               & 52.94    & 68.08          & 85.30         & 63.69        \\ \midrule
$h(\mathbf{x}^{new})$  & \multicolumn{1}{c|}{$\mathbf{x}^{old}$ }   & \multicolumn{1}{c|}{Base adapter \cite{FCT_side}}                     & 27.86               & 8.48   & 23.08         & 20.15        & 20.03        \\
$h(\mathbf{x}^{new})$  & \multicolumn{1}{c|}{$\mathbf{x}^{old}$ }   & \multicolumn{1}{c|}{EC-style adapter \cite{FCT_EC}}                     & 26.99               & 8.46    & 24.63        & 22.58       & 20.57     \\
$h(\mathbf{x}^{new})$  & \multicolumn{1}{c|}{$\mathbf{x}^{old}$ }   & \multicolumn{1}{c|}{CL-based adapter \cite{seo2025metric}}                & 39.80               & 27.42   & 44.60          & 51.65         & 39.27        \\
\rowcolor{gray!20} $h(\mathbf{x}^{new})$  & \multicolumn{1}{c|}{$\mathbf{x}^{old}$ }   & \multicolumn{1}{c|}{\ours}                      & \textbf{47.72}    & \textbf{37.11}   & \textbf{56.93}          & \textbf{63.18}        & \textbf{49.56}     \\ \midrule

\multicolumn{8}{c}{$\mathbf{x}^{old}$: E5-V \cite{e5v}, $\mathbf{x}^{new}$: UniME-7B \cite{unime}}                                                                                                                                                        \\ \midrule
$\mathbf{x}^{old}$      & \multicolumn{1}{c|}{$\mathbf{x}^{old}$ }   & \multicolumn{1}{c|}{-}                                & 45.54               & 12.49    & 47.98          & 54.88         & 38.21        \\
$\mathbf{x}^{new}$    & \multicolumn{1}{c|}{$\mathbf{x}^{new}$ }   & \multicolumn{1}{c|}{-}                                & 60.54               & 52.94    & 68.08          & 85.30         & 63.69        \\ \midrule
$h(\mathbf{x}^{new})$  & \multicolumn{1}{c|}{$\mathbf{x}^{old}$ }   & \multicolumn{1}{c|}{Base adapter \cite{FCT_side}}                     & 42.58               & 29.99    & 51.24          & 56.52         & 43.51        \\
$h(\mathbf{x}^{new})$  & \multicolumn{1}{c|}{$\mathbf{x}^{old}$ }   & \multicolumn{1}{c|}{EC-style adapter \cite{FCT_EC}}                & 43.23               & 30.91    & 51.72          & 56.83         & 44.14        \\
$h(\mathbf{x}^{new})$  & \multicolumn{1}{c|}{$\mathbf{x}^{old}$ }   & \multicolumn{1}{c|}{CL-based adapter \cite{seo2025metric} }                & 48.90               & 36.31    & 57.13          & 65.15         & 49.95        \\
\rowcolor{gray!20} $h(\mathbf{x}^{new})$  & \multicolumn{1}{c|}{$\mathbf{x}^{old}$ }   & \multicolumn{1}{c|}{\ours}                      & \textbf{50.58}    & \textbf{37.95}     & \textbf{57.76}          & \textbf{67.03 }        & \textbf{51.30}    \\ \bottomrule

\end{tabular}
    
    }
   \label{tables/mmeb_table}
\end{table}
\begin{equation} \label{eq:cl_loss}
\mathcal{L}_{CL} = -\sum_{i=1}^{N} \log \frac{\exp(\text{sim}(h(\mathbf{x}_{i}^{new}), \mathbf{x}_{i}^{old})/\tau)}{\sum_{j=1}^{N} \exp(\text{sim}(h(\mathbf{x}_{i}^{new}), \mathbf{x}_{j}^{old})/\tau)},
\end{equation}
where $\text{sim}(\cdot, \cdot)$ denotes cosine similarity and $\tau$ is the temperature parameter (set to 0.07). This loss maximizes similarity between the projected new embedding $h(\mathbf{x}_{i}^{new})$ and its corresponding old embedding $\mathbf{x}_{i}^{old}$ (positive pair), while minimizing similarity with other old embeddings (negatives). 
Note that our proposed regularization techniques can also be combined with this contrastive base adapter. We include these results in \cref{tables/analyses_cl}. 
\end{itemize}

\begin{table}[t!]
    \centering
     \caption{\textbf{Compatibility results in visual document and video benchmarks.} ViDore-V1 consists of 10 sub-datasets and is evaluated using nDCG@5, while video benchmarks are evaluated using Recall@1.}
     \resizebox{\columnwidth}{!}{
    \setlength{\tabcolsep}{3pt}
    \begin{tabular}{ccc|c|ccc|c}
\midrule
\multicolumn{2}{c|}{\textbf{Case}}        & \multirow{2}{*}{\textbf{Method}} & \textbf{Visual document}                                     & \multicolumn{3}{c|}{\textbf{Video}} & \multirow{2}{*}{\textbf{Average}} \\ \cline{4-7}
query     & \multicolumn{1}{c|}{database} &                                  & \begin{tabular}[c]{@{}c@{}}ViDoRe-V1 \cite{vidore} \\ (10) AVG\end{tabular} & MSR-VTT \cite{msrvtt}   & MSVD \cite{msvd}   & VATEX \cite{vatex}  &                                   \\ \midrule
\multicolumn{8}{c}{$\mathbf{x}^{old}$: CLIP ViT-L/14 \cite{openclip}, $\mathbf{x}^{new}$: GME-7B \cite{gme}}                                                                                                                                                                                   \\ \midrule
$\mathbf{x}^{old}$    & \multicolumn{1}{c|}{$\mathbf{x}^{old}$}   & -                                & 28.30                                                         & 28.60       & 36.27    & 20.43     & 28.40                             \\
$\mathbf{x}^{new}$    & \multicolumn{1}{c|}{$\mathbf{x}^{new}$}   & -                                & 89.40                                                        & 37.40       & 52.54    & 27.00     & 51.59                             \\ \midrule
$h(\mathbf{x}^{new})$& \multicolumn{1}{c|}{$\mathbf{x}^{old}$}   &  Base adapter \cite{FCT_side}                   & 22.04                                                       & 19.90   & 29.36     & 11.43       & 20.68                            \\

$h(\mathbf{x}^{new})$& \multicolumn{1}{c|}{$\mathbf{x}^{old}$}   &  EC-style adapter \cite{FCT_EC}                  & 24.01                                                      & 20.90    & 28.81     & 12.08      & 21.45 \\     
 $h(\mathbf{x}^{new})$ & \multicolumn{1}{c|}{$\mathbf{x}^{old}$}  &  CL-based adapter \cite{seo2025metric}                   & 38.47                                                              & 30.30   & 41.04     & 19.47     &32.32                \\
\rowcolor{gray!20} $h(\mathbf{x}^{new})$& \multicolumn{1}{c|}{$\mathbf{x}^{old}$}   & \ours                        & \textbf{45.35}                                                         & \textbf{33.60}       & \textbf{43.88}    & \textbf{22.18}     & \textbf{36.25}                      \\ \bottomrule
\end{tabular}
    
    }
    \label{tables/doc_video}
\end{table}

\begin{table}[t!]
    \centering
     \caption{\textbf{Ablation study on proposed components.} }
     \resizebox{0.9\columnwidth}{!}{
    \setlength{\tabcolsep}{3pt}
    \begin{tabular}{cc|l|ccc}
\midrule
\multicolumn{2}{c|}{\textbf{Case}}   & \multicolumn{1}{c|}{\textbf{Setting}}
    & \multirow{2}{*}{\textbf{I2T AVG}} & \multirow{2}{*}{\textbf{T2I AVG}} & \multirow{2}{*}{\textbf{MMEB AVG}}  \\
query     & database &                                                &                                       &                                                                  \\ \midrule
\multicolumn{6}{c}{$\mathbf{x}^{old}$: CLIP ViT-L/14 \cite{openclip}, $\mathbf{x}^{new}$: UniME-7B \cite{unime}}                                                                                                                                                                                                                                                   \\ \midrule

\rowcolor{gray!20} $h(\mathbf{x}^{new})$  & $\mathbf{x}^{old}$  & \ours & 77.76                                         & 63.07                                & 49.56  \\  \midrule

$h(\mathbf{x}^{new})$  & $\mathbf{x}^{old}$  &  w/o Point-wise preservation & 77.11                                           & 63.22                                  & 47.43                                                      \\
$h(\mathbf{x}^{new})$  & $\mathbf{x}^{old}$  & w/o Distance-wise preservation & 71.74                                          & 58.08                               & 47.81                                              \\
$h(\mathbf{x}^{new})$  & $\mathbf{x}^{old}$  & w/o Angle-wise preservation & 76.83                                       & 62.26                                & 41.55                                                   \\
$h(\mathbf{x}^{new})$  & $\mathbf{x}^{old}$ & w/o Focal re-weighting  & 77.43                                        & 62.98                                & 46.61    
\\ \bottomrule
\end{tabular}
    
    }
    \label{tables/ablation_main}
\end{table}

We omit XBT~\cite{BCT_XBT} from the main tables for two reasons.
First, XBT requires LoRA-tuning~\cite{lora} of the new embedding model, making it an unfair comparison to our adapter-only setting.
Second, as detailed in \cref{sec:appendix_xbt}, XBT performs worse in our MLLM compatibility settings, even compared to other adapter-only baselines, likely because it disrupts the pre-trained representations of well-trained MLLMs.
Additional analyses and results are provided in S.M.

\subsection{Main results} 
\cref{fig:results_histogram} provides an overview of backward compatibility performance across three representative model upgrade scenarios in image-text-based benchmarks; the y-axis shows the performance gap relative to the old model, with the black dashed line at zero indicating the old model's performance. 
Overall, \ours (shown in red) consistently improves over the old model and all baseline adapters on both standard I2T/T2I retrieval and the more challenging MMEB benchmark, with average (across scenarios) gains of $+5.43\%$ on I2T, $+7.63\%$ on T2I, and $+10.53\%$ on MMEB.
In contrast, existing baselines often fail to preserve backward compatibility (\ie, they underperform the old model itself).
Specifically, on standard I2T/T2I retrieval (\cref{fig:results_histogram} and \cref{tables/i2t_t2i_table}), CL-based adapters typically degrade I2T performance (\eg, in the CLIP$\rightarrow$GME and CLIP$\rightarrow$UniME scenarios), while $L2$-based adapters (Base and EC-style) often degrade T2I performance.
On the other hand, \ours consistently improves performance on both tasks, achieving substantial gains in the CLIP$\rightarrow$UniME scenario ($+7.96\%$ on I2T and $+10.6\%$ on T2I on average).
On the MMEB benchmark (\cref{fig:results_histogram} and \cref{tables/mmeb_table}), CL-based adapters generally outperform $L2$-based ones, consistent with their stronger T2I performance in \cref{tables/i2t_t2i_table}, as MMEB often involves text-heavy queries. However, they still fail to maintain backward compatibility in some scenarios (\eg, CLIP$\rightarrow$UniME).
In contrast, \ours consistently achieves the strongest performance across upgrade scenarios, with particularly large gains in the CLIP$\rightarrow$UniME setting (averaging $+9.30\%$ over the old model), showing robust backward compatibility and out-of-domain generalization.

We further report results on text-to-visual-document and text-to-video retrieval in \cref{tables/doc_video}.
\ours achieves substantial gains on both tasks, demonstrating strong generalization ($+17.05\%$ on visual documents and $+4.80\%$ on videos).
In this setting, $L2$-based adapters perform significantly worse than on other benchmarks; we therefore adopt a contrastive learning (CL) objective as the base adapter.
\cref{tables/analyses_cl} provides additional evidence that our method remains effective when combined with CL-based adapters, confirming the modularity and versatility of our approach. Further details are provided in S.M.

\begin{table}[t]
\centering
\begin{minipage}[t]{0.45\columnwidth}
    \centering
    \captionof{table}{\textbf{Analyses on preservation loss.}}    
    \label{tables:analyses_1}
    \resizebox{\linewidth}{!}{
        \setlength{\tabcolsep}{3pt}
        
\begin{tabular}{ccc|cc}
\midrule
\multicolumn{2}{c|}{\textbf{Case}}         & \multirow{2}{*}{\textbf{Method}}  & \multicolumn{2}{c}{\textbf{Average}} \\ \cline{4-5} 
query      & \multicolumn{1}{c|}{database} &                                   & I2T R@1                  & T2I R@1   \\ \midrule
\multicolumn{5}{c}{$\mathbf{x}^{old}$: E5-V-7B \cite{e5v}, $\mathbf{x}^{new}$: UniME-7B \cite{unime}}         \\ \midrule
$\mathbf{x}^{old}$     & \multicolumn{1}{c|}{$\mathbf{x}^{old}$}   & -                                 & 78.08                    & 72.41     \\
$\mathbf{x}^{new}$     & \multicolumn{1}{c|}{$\mathbf{x}^{new}$}   & -                                  & 86.52                    & 77.15     \\ \midrule
$h(\mathbf{x}^{new})$  & \multicolumn{1}{c|}{$\mathbf{x}^{old}$}   & Base adapter                      & 72.93                    & 72.11     \\
\rowcolor{gray!20} $h(\mathbf{x}^{new})$  & \multicolumn{1}{c|}{$\mathbf{x}^{old}$}   & \ours           & 82.03                    & 75.27      \\ \midrule

$h(\mathbf{x}^{new})$  & \multicolumn{1}{c|}{$h(\mathbf{x}^{new})$}   & Base adapter                       & 76.88                        & 70.71   \\
\rowcolor{gray!20} $h(\mathbf{x}^{new})$  & \multicolumn{1}{c|}{$h(\mathbf{x}^{new})$}   & \ours            & 86.30                      & 77.11    \\ \bottomrule

\end{tabular}

    }
\end{minipage}
\hfill
\begin{minipage}[t]{0.45\columnwidth}
    \centering
     \captionsetup{type=figure}
    \captionof{figure}{\textbf{The impact of lambda.} We omit the case without the base loss.}
    \label{fig:preservation_graph}
    \includegraphics[width=\linewidth]{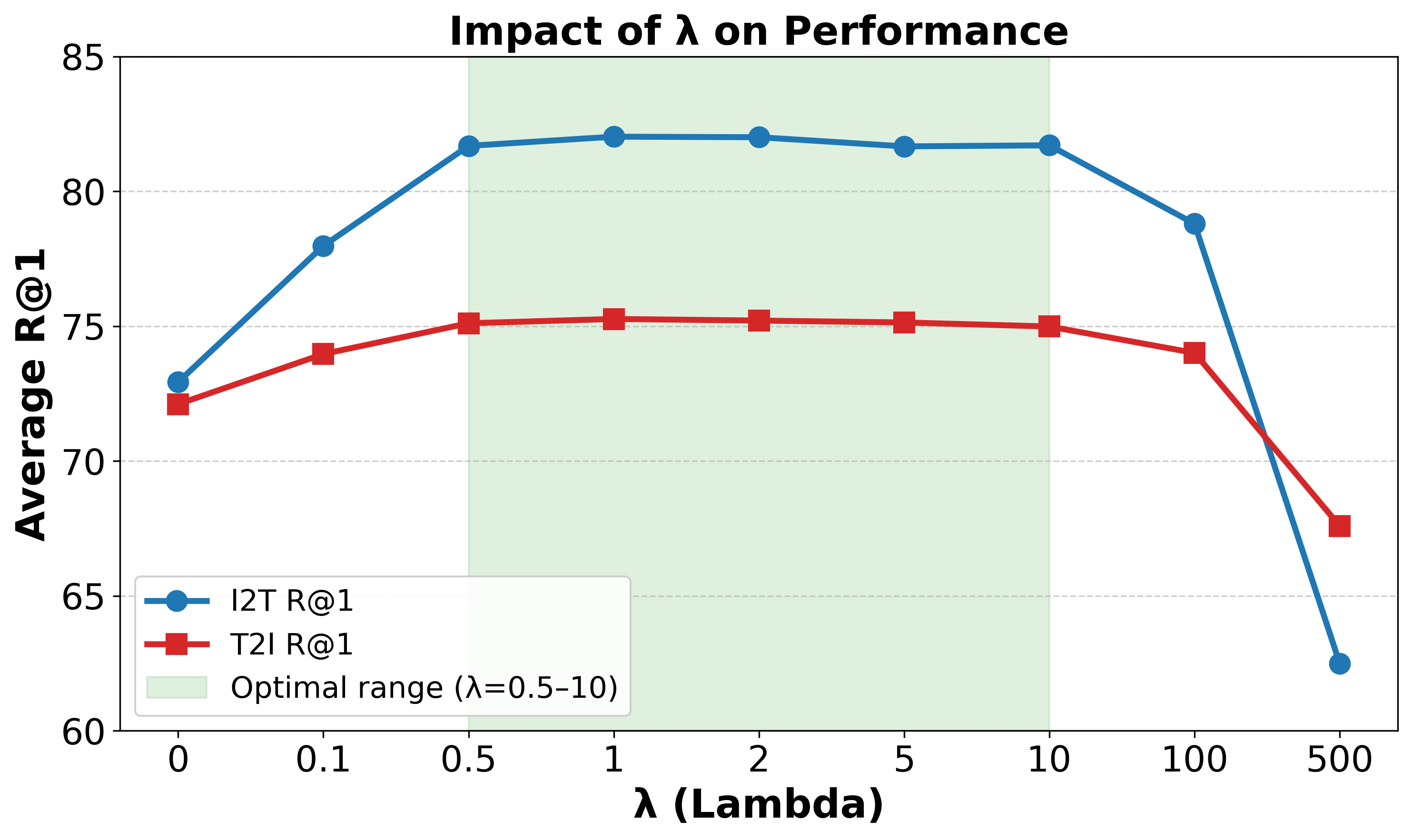}
\end{minipage}
\end{table}

\begin{table}[t!]
    \centering
     \caption{\textbf{Analyses on adapters and training sets.} Details follow \cref{tables/i2t_t2i_table}. }
     \resizebox{\columnwidth}{!}{
    \setlength{\tabcolsep}{3pt}
    
\begin{tabular}{ccc|cccccccc}
\midrule
\multicolumn{2}{c|}{\textbf{Case}}         & \multirow{2}{*}{\textbf{Variants}} & \multicolumn{2}{c}{\textbf{COCO}} & \multicolumn{2}{c}{\textbf{Flickr30K}} & \multicolumn{2}{c|}{\textbf{Urban1K}} & \multicolumn{2}{c}{\textbf{Average}} \\ \cline{4-11} 
query      & \multicolumn{1}{c|}{database} &                                  & I2T R@1         & T2I R@1         & I2T R@1          & T2I R@1          & I2T R@1 & \multicolumn{1}{c|}{T2I R@1} & I2T R@1                  & T2I R@1   \\ \midrule
\multicolumn{11}{c}{$\mathbf{x}^{old}$: E5-V \cite{e5v}, $\mathbf{x}^{new}$: UniME-7B \cite{unime}}                                                                                                                                                                                        \\ \midrule
\rowcolor{gray!20} $h(\mathbf{x}^{new})$  & \multicolumn{1}{c|}{$\mathbf{x}^{old}$}   & Original                     & 65.28          & 52.61           & 90.60            & 81.10           & 90.20   & \multicolumn{1}{c|}{92.10}   & 82.03                    & 75.27  \\ \midrule
$h(\mathbf{x}^{new})$  & \multicolumn{1}{c|}{$\mathbf{x}^{old}$}   & Adapter $\times 2$                      & 65.72               & 52.70               & 90.20                & 81.08               & 91.40       & \multicolumn{1}{c|}{92.30}       & 82.44                      & 75.36   \\

$h(\mathbf{x}^{new})$  & \multicolumn{1}{c|}{$\mathbf{x}^{old}$}   & Adapter $\times 4$                      & 65.92               & 52.91               & 90.80                & 81.46               & 91.10       & \multicolumn{1}{c|}{92.20}       & 82.61                       & 75.52   \\
$h(\mathbf{x}^{new})$  & \multicolumn{1}{c|}{$\mathbf{x}^{old}$}   & Adapter $\times 8$                    & 66.42               & 52.94               & 90.70                & 81.24                & 91.40       & \multicolumn{1}{c|}{92.10}       & 82.84                     & 75.43   \\ \midrule

$h(\mathbf{x}^{new})$  & \multicolumn{1}{c|}{$\mathbf{x}^{old}$}  &  SBU (1M pairs)               & 66.22              & 52.82            & 90.00              & 81.12               & 91.50      & \multicolumn{1}{c|}{92.00}       & 82.57                       & 75.31   \\ 
$h(\mathbf{x}^{new})$  & \multicolumn{1}{c|}{$\mathbf{x}^{old}$}   &  CC3M+SBU (4M pairs)                & 66.02               & 52.66              & 90.60               & 81.44               & 91.00       & \multicolumn{1}{c|}{91.60}       & 82.54                      & 75.23   \\ 
\bottomrule

\end{tabular}
    
    }
    \label{tables/analyses_2}
\end{table}

\subsection{Ablation studies}
We evaluate the effectiveness of our proposed components in \cref{tables/ablation_main}, including the multi-level preservation losses (point-, distance-, and angle-wise) and the focal re-weighting strategy.
All adapters are trained on LLaVA-LCS under the same setup as in \cref{tables/i2t_t2i_table,tables/mmeb_table}. We focus on the CLIP ViT-L/14 (old) $\rightarrow$ UniME (new) scenario, where the first row corresponds to the full model.
Removing any component consistently degrades performance across metrics, indicating complementary contributions.
In particular, removing the ``distance loss'' degrades standard I2T/T2I performance, whereas removing other components more strongly affects MMEB metrics.
Additional results in \cref{tables/ablation_suppl_preserve} and \cref{tables/ablation_suppl_focal} analyze the standalone impact of each component.
Each preservation loss yields positive gains individually and complementary improvements when combined (\cref{tables/ablation_suppl_preserve}).
Focal re-weighting consistently improves performance across different loss combinations (\cref{tables/ablation_suppl_focal}), demonstrating its robustness.

\subsection{Analyses} \label{sec:analyses}
\noindent{\textbf{[Impact of preservation loss]}} 
We validate our core hypothesis: whether our method preserves the representational quality of the new embeddings. To this end, we evaluate retrieval when both the query and database use the projected embeddings $h(\mathbf{x}^{new})$ (rows 5-6), instead of using the original old embeddings in the database as in the standard BCT setting (rows 3-4).
If the similarity structure of the new embeddings is preserved after projection, the resulting performance should remain close to that of the new model and considerably exceed that of the old model.
As shown in the last row of \cref{tables:analyses_1}, the projected embeddings from ours (row 6) achieve performance within 0.2 points of the new embedding model (row 2), indicating that \ours preserves retrieval quality in the projected space and thus enables effective backward compatibility (row 4).

Next, we analyze the trade-off between projection and preservation by varying $\lambda$. 
When $\lambda=\infty$, the base projection loss is removed, leading to near-zero performance, as expected.
Performance peaks around $\lambda\in[0.5,10]$, while overly large values ($\lambda > 100$) degrade results by over-emphasizing preservation. Our method remains stable across a broad moderate range, indicating robustness to $\lambda$. Sensitivity to other hyperparameters, including the relative weights among preservation losses and the focal parameter $\gamma$, is reported in \cref{tables/ablation_hyperparameter}.

\noindent{\textbf{[Adapter sizes and dataset scales]}} Larger adapters yield slightly better results, but the overall trends remain consistent. Increasing the training data from 558k pairs (LLaVA-LCS) to 4M pairs (CC3M+SBU) provides only marginal gains, suggesting that our adapter design and a moderate-scale dataset are already sufficient to achieve strong and robust performance across benchmarks.

\noindent{\textbf{[Computational overhead] }} 
We first measure the inference cost of our adapter on a single A100 (40GB) with a batch size of 32. In the CLIP ViT-L/14 \cite{openclip} to UniME (7B) \cite{unime} setting, our adapter adds only 0.25 ms of latency per batch (0.01\% latency to the original backbone forward pass).
Training is also efficient, requiring less than 10 minutes per epoch on a 1M-sample dataset, whereas LoRA-tuning a 7B model typically takes over a day per epoch even on H100 GPUs.
Although we use 50 epochs by default, 10 epochs already achieve comparable performance (see S.M.), enabling complete training in 1–2 hours on 8 A100 GPUs. In contrast, backfilling even a moderate 5M-scale database takes about one day.

\section{Conclusion}
\label{sec:conclusion}

We propose a practical adapter-only approach for achieving backward compatibility with MLLMs.
We devise multi-level preservation losses with focal re-weighting to retain the representational quality of new embeddings after projection.
Experiments across diverse modalities demonstrate strong backward compatibility with negligible latency and efficient training.

\bibliographystyle{splncs04}
\bibliography{11_references}

\clearpage  

\clearpage 
\appendix
\setcounter{page}{1}
\renewcommand\thesection{\Alph{section}}
\renewcommand{\theequation}{S\arabic{equation}}
\setcounter{section}{0}
\setcounter{table}{0}

\onecolumn

{
    \centering
    \Large
    \textbf{Multi-modal Knowledge Preserving Adapter for Embedding Backward Compatibility} 
\\
    \vspace{0.5em}Supplementary Material \\
    \vspace{1.0em}
}

\renewcommand{\thetable}{S\arabic{table}}

\section{Details on loss functions} \label{sec:appendix_loss}
\subsection{Details on preservation loss} \label{sec:appendix_preservation_loss}

As described in \cref{sec:method}, both the distance-wise and angle-wise preservation losses are composed of two parts: a global term and a local term. Each preservation loss is defined as the sum of these two components, as defined in \cref{eq:distance_sum,eq:angle_sum}.
In the global distance-wise preservation loss (\cref{eq:global_distance}), index $i$ denotes an anchor sample from the mini-batch, and $\mathbf{x}_j$ denotes another sample randomly drawn from the same mini-batch. In contrast, the local distance-wise preservation loss (\cref{eq:local_distance}) considers the $m$ nearest neighbors of each anchor and enforces the preservation of their relative distance relationships. We select nearest neighbors in the new embedding space, since it preserves better semantic structure, which aligns with our core motivation. We set $m=100$ in all experiments.

\begin{equation} \label{eq:distance_sum}
\mathcal{L}_{distance} = \mathcal{L}_{{distance}, global} + \mathcal{L}_{{distance},local}
\end{equation}

\begin{equation} \label{eq:global_distance}
    \mathcal{L}_{{distance}, global} = \sum_{(i,j) \in \mathcal{P}_{global}} \left| Dist(h(\mathbf{x}_{i}^{new}), h(\mathbf{x}_{j}^{new})) - Dist(\mathbf{x}_{i}^{new}, \mathbf{x}_{j}^{new}) \right|,
\end{equation}

\begin{equation} \label{eq:local_distance}
\mathcal{L}_{{distance},local} = \sum_{i} \sum_{j \in \mathcal{N}_m(i)} \left| Dist(h(\mathbf{x}_{i}^{new}), h(\mathbf{x}_{j}^{new})) - Dist(\mathbf{x}_{i}^{new}, \mathbf{x}_{j}^{new}) \right|,
\end{equation}

Similar to the distance-wise loss, the global angle-wise preservation loss in \cref{eq:global_angle} follows the same indexing scheme: $j$ denotes the anchor sample, while $i$ and $k$ denote randomly sampled samples from the mini-batch that are used to form angular relationships. The local angle-wise loss, defined in \cref{eq:local_angle}, considers the $m$ nearest neighbors of each anchor. To control computational cost, we limit the number of sampled pairs $(i,k)$ per anchor to 32 for both the global and local angle-wise losses.

\begin{equation}  \label{eq:angle_sum}
\mathcal{L}_{angle} = \mathcal{L}_{angle, global} + \mathcal{L}_{angle,local}
\end{equation}

\begin{equation} \label{eq:global_angle}
\mathcal{L}_{angle,global} = \sum_{(i,j,k) \in \mathcal{T}_{global}} \left| \psi_A(h(\mathbf{x}_{i}^{new}), h(\mathbf{x}_{j}^{new}), h(\mathbf{x}_{k}^{new})) - \psi_A(\mathbf{x}_{i}^{new}, \mathbf{x}_{j}^{new}, \mathbf{x}_{k}^{new}) \right|,
\end{equation}

\begin{equation}  \label{eq:local_angle}
\mathcal{L}_{angle,local} = \sum_{j} \sum_{i,k \in \mathcal{N}_m(j)} \left| \psi_A(h(\mathbf{x}_{i}^{new}), h(\mathbf{x}_{j}^{new}), h(\mathbf{x}_{k}^{new})) - \psi_A(\mathbf{x}_{i}^{new}, \mathbf{x}_{j}^{new}, \mathbf{x}_{k}^{new}) \right|,
\end{equation}

\subsection{Details on focal re-weighting strategy} 
Similar to the point-wise preservation loss in \cref{eq:point_focal_loss}, we compute the focal weights for the distance-wise and angle-wise preservation losses based on their respective preservation errors, as defined below:

\begin{equation} \label{eq:distance_focal_loss}
\begin{split}
\mathcal{L}_{distance}^{focal} &= \frac{1}{|\mathcal{P}|}\sum_{(i,j) \in \mathcal{P}} w_{ij} \Delta_{ij}, \\
\text{where} \quad & \Delta_{ij} = \left|\text{Dist}(h(\mathbf{x}_{i}^{new}), h(\mathbf{x}_{j}^{new})) - \text{Dist}(\mathbf{x}_{i}^{new}, \mathbf{x}_{j}^{new})\right|, \\
& w_{ij} = \left(\frac{\Delta_{ij}}{\bar{\Delta}_{distance}}\right)^{\gamma}, \quad \bar{\Delta}_{distance} = \frac{1}{|\mathcal{P}|}\sum_{(i,j) \in \mathcal{P}} \Delta_{ij}.
\end{split}
\end{equation}

\begin{equation} \label{eq:angle_focal_loss}
\begin{split}
\mathcal{L}_{angle}^{focal} &= \frac{1}{|\mathcal{T}|}\sum_{(i,j,k) \in \mathcal{T}} w_{ijk} \Delta_{ijk}, \\
\text{where} \quad & \Delta_{ijk} = \left|\psi_A(h(\mathbf{x}_{i}^{new}), h(\mathbf{x}_{j}^{new}), h(\mathbf{x}_{k}^{new})) - \psi_A(\mathbf{x}_{i}^{new}, \mathbf{x}_{j}^{new}, \mathbf{x}_{k}^{new})\right|, \\
& w_{ijk} = \left(\frac{\Delta_{ijk}}{\bar{\Delta}_{angle}}\right)^{\gamma}, \quad \bar{\Delta}_{angle} = \frac{1}{|\mathcal{T}|}\sum_{(i,j,k) \in \mathcal{T}} \Delta_{ijk}.
\end{split}
\end{equation}

\section{Additional details and results}
\subsection{Additional implementation details} \label{sec:appendix_details}
We use Pytorch 2.4 with CUDA 12.5.
We use CLIP ViT-L/14 with 768-dimensional embeddings. E5-V and UniME share the LLaVA-1.6 (7B) backbone \cite{llavanext} and produce 4096-dimensional embeddings, whereas GME (7B) is based on Qwen2-VL \cite{qwen2-vl} and produces 3584-dimensional embeddings.

\subsection{More results on main ablations} \label{sec:appendix_focal}
We further analyze the standalone impact of each preservation loss and the focal re-weighting strategy in \cref{tables/ablation_suppl_preserve} and \cref{tables/ablation_suppl_focal}. As shown in \cref{tables/ablation_suppl_preserve}, each component provides distinct and complementary benefits. The point-wise preservation loss mainly improves MMEB, while the distance-wise preservation loss primarily boosts I2T and T2I performance, and the angle-wise preservation loss yields more balanced improvements across metrics. Combining all three preservation losses consistently achieves the best overall performance, highlighting their complementary nature.
In \cref{tables/ablation_suppl_focal}, we observe that across various combinations of preservation loss variants, incorporating the focal re-weighting strategy enhances performance on average (with an overall improvement of +2.11\%), demonstrating its effectiveness.

\begin{table}[!ht]
    \centering
     \caption{\textbf{Standalone analysis on each preservation loss. } Details are the same as in \cref{tables/ablation_main}.}
     \resizebox{0.8\columnwidth}{!}{
    \setlength{\tabcolsep}{3pt}
    
\begin{tabular}{cccccccc}
\midrule
\multicolumn{2}{c|}{\textbf{Case}}        & \multicolumn{3}{c|}{\textbf{Ours}}              & \multirow{2}{*}{\textbf{ I2T AVG}} & \multirow{2}{*}{\textbf{ T2I AVG}} & \multicolumn{1}{c|}{\multirow{2}{*}{\textbf{MMEB AVG}}}  \\
query     & \multicolumn{1}{c|}{database} & Point & Distance    & \multicolumn{1}{c|}{Angle} &                                                &                                       & \multicolumn{1}{c|}{}                                                                 \\ \midrule
\multicolumn{8}{c}{$\mathbf{x}^{old}$: CLIP ViT-L/14 \cite{clip}, $\mathbf{x}^{new}$: UniME-7B \cite{unime}}                                                                                                                                                                                                                                                   \\ \midrule

$h(\mathbf{x}^{new})$  & \multicolumn{1}{c|}{$\mathbf{x}^{old}$ }   & -   & -          & \multicolumn{1}{c|}{-}     & 63.35                                           & 56.39                                  & \multicolumn{1}{c|}{20.03}                                                        \\ \midrule
$h(\mathbf{x}^{new})$  & \multicolumn{1}{c|}{$\mathbf{x}^{old}$ }   & O   & -         & \multicolumn{1}{c|}{-}     & 67.01                                           & 58.04                                 & \multicolumn{1}{c|}{34.95}                                                   \\
$h(\mathbf{x}^{new})$  & \multicolumn{1}{c|}{$\mathbf{x}^{old}$ }   & -   & O        & \multicolumn{1}{c|}{-}     & 72.70                                           & 58.73                                 & \multicolumn{1}{c|}{28.60}                                                   \\
$h(\mathbf{x}^{new})$  & \multicolumn{1}{c|}{$\mathbf{x}^{old}$ }   & -   & -        & \multicolumn{1}{c|}{O}     & 72.40                                           & 58.00                                 & \multicolumn{1}{c|}{33.22}                                                   \\ \midrule
$h(\mathbf{x}^{new})$  & \multicolumn{1}{c|}{$\mathbf{x}^{old}$ }   & O   & O       & \multicolumn{1}{c|}{-}     & 76.15                                         & 60.76                                 & \multicolumn{1}{c|}{32.94}                                                    \\
 $h(\mathbf{x}^{new})$  & \multicolumn{1}{c|}{$\mathbf{x}^{old}$ }   & O   &  -        &\multicolumn{1}{c|}{O}     & 72.73                                          & 58.77                                & \multicolumn{1}{c|}{45.43}       \\

 $h(\mathbf{x}^{new})$  & \multicolumn{1}{c|}{$\mathbf{x}^{old}$ }   & -  &  O        &\multicolumn{1}{c|}{O}     & 75.88                                          & 62.21                               & \multicolumn{1}{c|}{39.25}       \\
 $h(\mathbf{x}^{new})$  & \multicolumn{1}{c|}{$\mathbf{x}^{old}$ }   & O   & O         & \multicolumn{1}{c|}{O}      & 77.43                                         & 62.98                               & \multicolumn{1}{c|}{46.61}       
\\ \bottomrule
\end{tabular}
    
    }
    \label{tables/ablation_suppl_preserve}
\end{table}

\begin{table}[h!]
    \centering
     \caption{\textbf{Standalone analysis of focal reweighting strategy.} Details are the same as in \cref{tables/ablation_main}.}
     \resizebox{0.8\columnwidth}{!}{
    \setlength{\tabcolsep}{3pt}
    
\begin{tabular}{cccccc}
\midrule
\multicolumn{2}{c|}{\textbf{Case}}        & \multicolumn{1}{c|}{\textbf{Ours}}              & \multirow{2}{*}{\textbf{ I2T AVG}} & \multirow{2}{*}{\textbf{ T2I AVG}} & \multicolumn{1}{c|}{\multirow{2}{*}{\textbf{MMEB AVG}}}  \\
query     & \multicolumn{1}{c|}{database} & Variants &                                                &                                       & \multicolumn{1}{c|}{}                                                                 \\ \midrule
\multicolumn{6}{c}{$\mathbf{x}^{old}$: CLIP ViT-L/14 \cite{clip}, $\mathbf{x}^{new}$: UniME-7B \cite{unime}}                                                                                                                                                                                                                                                   \\ \midrule

$h(\mathbf{x}^{new})$  & \multicolumn{1}{c|}{$\mathbf{x}^{old}$ }           & \multicolumn{1}{c|}{Point+Distance}     & 76.15                                     & 60.76                                & \multicolumn{1}{c|}{32.94}                                                        \\
$h(\mathbf{x}^{new})$  & \multicolumn{1}{c|}{$\mathbf{x}^{old}$ }           & \multicolumn{1}{c|}{Point+Distance+\textbf{Focal}}     & 76.83                                          & 62.26                                & \multicolumn{1}{c|}{41.55}                                                   \\\midrule
$h(\mathbf{x}^{new})$  & \multicolumn{1}{c|}{$\mathbf{x}^{old}$ }           & \multicolumn{1}{c|}{Point+Angle}     & 72.73                                           & 58.77                                 & \multicolumn{1}{c|}{45.43}                                                   \\
$h(\mathbf{x}^{new})$  & \multicolumn{1}{c|}{$\mathbf{x}^{old}$ }           & \multicolumn{1}{c|}{Point+Angle+\textbf{Focal}}     & 71.74                                           & 58.08                                 & \multicolumn{1}{c|}{47.81}                                                   \\ \midrule
$h(\mathbf{x}^{new})$  & \multicolumn{1}{c|}{$\mathbf{x}^{old}$ }          & \multicolumn{1}{c|}{Distance+Angle}     & 75.88                                         & 62.21                                 & \multicolumn{1}{c|}{39.25}                                                    \\
 $h(\mathbf{x}^{new})$  & \multicolumn{1}{c|}{$\mathbf{x}^{old}$ }         &\multicolumn{1}{c|}{Distance+Angle+\textbf{Focal}}     & 77.11                                          & 63.22                                & \multicolumn{1}{c|}{47.43}       \\ \midrule

 $h(\mathbf{x}^{new})$  & \multicolumn{1}{c|}{$\mathbf{x}^{old}$ }        &\multicolumn{1}{c|}{Point+Distance+Angle}    & 77.43                                          & 62.98                               & \multicolumn{1}{c|}{46.61}            \\
 $h(\mathbf{x}^{new})$  & \multicolumn{1}{c|}{$\mathbf{x}^{old}$ }           & \multicolumn{1}{c|}{Point+Distance+Angle+\textbf{Focal}}      
  & 77.76                                          & 63.07                               & \multicolumn{1}{c|}{49.56} 
\\ \bottomrule
\end{tabular}
    
    }
    \label{tables/ablation_suppl_focal}
\end{table}

\subsection{Batch size ablation} \label{sec:appendix_batch}
We evaluate our method across various batch sizes and find that performance remains stable. While the CL-based method benefits slightly from larger batch sizes by leveraging more negative samples in contrastive learning, the overall impact on performance is marginal.

\begin{table}[h!]
    \centering
     \caption{\textbf{Analyses on batch size}. Details are the same as in \cref{tables/i2t_t2i_table}. }
     \resizebox{\columnwidth}{!}{
    \setlength{\tabcolsep}{3pt}
    
\begin{tabular}{ccc|cccccccc}
\midrule
\multicolumn{2}{c|}{\textbf{Case}}         & \multirow{2}{*}{\textbf{Batch size}} & \multicolumn{2}{c}{\textbf{COCO}} & \multicolumn{2}{c}{\textbf{Flickr30K}} & \multicolumn{2}{c|}{\textbf{Urban1K}} & \multicolumn{2}{c}{\textbf{Average}} \\ \cline{4-11} 
query      & \multicolumn{1}{c|}{database} &                                  & I2T R@1         & T2I R@1         & I2T R@1          & T2I R@1          & I2T R@1 & \multicolumn{1}{c|}{T2I R@1} & I2T R@1                  & T2I R@1   \\ \midrule
\multicolumn{11}{c}{Batch 32}                                                                                                                                                                                        \\ \midrule

$h(\mathbf{x}^{new})$  & \multicolumn{1}{c|}{$\mathbf{x}^{old}$}   & Base adapter \cite{FCT_side}                     & 55.90           & 50.21           & 85.10            & 79.22            & 77.80   & \multicolumn{1}{c|}{86.90}   & 72.93                    & 72.11     \\
$h(\mathbf{x}^{new})$  & \multicolumn{1}{c|}{$\mathbf{x}^{old}$}   & EC-style adapter \cite{FCT_EC}                      & 57.30               & 50.32               & 85.90               & 79.52                & 78.10       & \multicolumn{1}{c|}{88.10}       & 73.77                        & 72.65      \\
$h(\mathbf{x}^{new})$  & \multicolumn{1}{c|}{$\mathbf{x}^{old}$}   & CL-based adapter \cite{seo2025metric}                     & 63.18               & 50.12               & 88.70                & 78.46                & 81.70       & \multicolumn{1}{c|}{88.70}       & 77.86                        & 72.43      \\
\rowcolor{gray!20} $h(\mathbf{x}^{new})$  & \multicolumn{1}{c|}{$\mathbf{x}^{old}$}   & \ours                        & 65.28          & 52.61           & 90.60            & 81.10           & 90.20   & \multicolumn{1}{c|}{92.10}   & 82.03                    & 75.27     \\  \midrule

\multicolumn{11}{c}{Batch 128}                    \\ \midrule
$h(\mathbf{x}^{new})$  & \multicolumn{1}{c|}{$\mathbf{x}^{old}$}   & Base adapter \cite{FCT_side}                     & 56.22          & 50.36        & 84.60         & 79.14          & 78.40  & \multicolumn{1}{c|}{87.90 }   &  73.07               & 72.47    \\
$h(\mathbf{x}^{new})$  & \multicolumn{1}{c|}{$\mathbf{x}^{old}$}   & EC-style adapter \cite{FCT_EC}                        & 58.54          & 50.74         & 86.60         & 79.62          & 79.80 & \multicolumn{1}{c|}{88.10}   & 74.98                & 72.82     \\
$h(\mathbf{x}^{new})$  & \multicolumn{1}{c|}{$\mathbf{x}^{old}$}   & CL-based adapter \cite{seo2025metric}                       & 63.56          & 50.04         & 88.60         & 78.50          & 80.60 & \multicolumn{1}{c|}{89.10}   &    77.59             & 72.55     \\
\rowcolor{gray!20} $h(\mathbf{x}^{new})$  & \multicolumn{1}{c|}{$\mathbf{x}^{old}$}   & \ours                               & 65.92               & 52.91               & 90.80                & 81.46               & 91.10       & \multicolumn{1}{c|}{92.20}       & 82.61                       & 75.52     \\
\midrule

\multicolumn{11}{c}{Batch 1024}                    \\ \midrule
$h(\mathbf{x}^{new})$  & \multicolumn{1}{c|}{$\mathbf{x}^{old}$}   & Base adapter \cite{FCT_side}                     & 56.32           & 50.52         & 85.20          & 79.46         & 78.60 & \multicolumn{1}{c|}{88.00}   & 73.37                & 72.66    \\
$h(\mathbf{x}^{new})$  & \multicolumn{1}{c|}{$\mathbf{x}^{old}$}   & EC-style adapter \cite{FCT_EC}                        & 58.66           & 50.95         & 86.10          & 79.46          & 79.40 & \multicolumn{1}{c|}{88.40}   & 74.72                & 72.93    \\
$h(\mathbf{x}^{new})$  & \multicolumn{1}{c|}{$\mathbf{x}^{old}$}   & CL-based adapter \cite{seo2025metric}                       & 63.88          & 50.15         & 89.20         & 78.70          & 81.70 & \multicolumn{1}{c|}{89.30}   & 78.26                & 72.72     \\
\rowcolor{gray!20} $h(\mathbf{x}^{new})$  & \multicolumn{1}{c|}{$\mathbf{x}^{old}$}   & \ours                        &       65.30    & 52.78       & 90.60          & 81.06          & 90.90 & \multicolumn{1}{c|}{91.60}   & 82.27               & 75.15     \\
\bottomrule

\end{tabular}
    
    }
    \label{tables/analyses_batch}
\end{table}

\subsection{Adapter size ablation} \label{sec:analyses_hidden}
As shown in \cref{tables/analyses_hidden}, increasing the hidden size yields marginal gains, suggesting that the current adapter capacity is sufficient for learning the mapping.
\begin{table}[h!]
    \centering
     \caption{\textbf{Analyses on hidden layers}. Details are the same as in \cref{tables/i2t_t2i_table}. }
     \resizebox{\columnwidth}{!}{
    \setlength{\tabcolsep}{3pt}
    
\begin{tabular}{ccc|cccccccc}
\midrule
\multicolumn{2}{c|}{\textbf{Case}}         & \multirow{2}{*}{\textbf{Method}} & \multicolumn{2}{c}{\textbf{COCO}} & \multicolumn{2}{c}{\textbf{Flickr30K}} & \multicolumn{2}{c|}{\textbf{Urban1K}} & \multicolumn{2}{c}{\textbf{Average}} \\ \cline{4-11} 
query      & \multicolumn{1}{c|}{database} &                                  & I2T R@1         & T2I R@1         & I2T R@1          & T2I R@1          & I2T R@1 & \multicolumn{1}{c|}{T2I R@1} & I2T R@1                  & T2I R@1   \\ \midrule
\multicolumn{11}{c}{Hidden $\times 1$}                                                                                                                                                                                        \\ \midrule

$h(\mathbf{x}^{new})$  & \multicolumn{1}{c|}{$\mathbf{x}^{old}$}   & Base adapter \cite{FCT_side}                     & 55.90           & 50.21           & 85.10            & 79.22            & 77.80   & \multicolumn{1}{c|}{86.90}   & 72.93                    & 72.11     \\
$h(\mathbf{x}^{new})$  & \multicolumn{1}{c|}{$\mathbf{x}^{old}$}   & EC-style adapter \cite{FCT_EC}                      & 57.30               & 50.32               & 85.90               & 79.52                & 78.10       & \multicolumn{1}{c|}{88.10}       & 73.77                        & 72.65      \\
$h(\mathbf{x}^{new})$  & \multicolumn{1}{c|}{$\mathbf{x}^{old}$}   & CL-based adapter \cite{seo2025metric}                     & 63.18               & 50.12               & 88.70                & 78.46                & 81.70       & \multicolumn{1}{c|}{88.70}       & 77.86                        & 72.43      \\
\rowcolor{gray!20} $h(\mathbf{x}^{new})$  & \multicolumn{1}{c|}{$\mathbf{x}^{old}$}   & \ours                      & 65.28          & 52.61           & 90.60            & 81.10           & 90.20   & \multicolumn{1}{c|}{92.10}   & 82.03                    & 75.27     \\  \midrule

\multicolumn{11}{c}{Hidden $\times 2$}                    \\ \midrule
$h(\mathbf{x}^{new})$  & \multicolumn{1}{c|}{$\mathbf{x}^{old}$}   & Base adapter \cite{FCT_side}                     & 56.22          & 50.36        & 84.60         & 79.14          & 78.40  & \multicolumn{1}{c|}{87.90 }   &  73.07               & 72.47    \\
$h(\mathbf{x}^{new})$  & \multicolumn{1}{c|}{$\mathbf{x}^{old}$}   & EC-style adapter \cite{FCT_EC}                        & 58.02          & 50.43         &86.30         & 79.34          & 80.00 & \multicolumn{1}{c|}{87.70}   & 74.77                & 72.49     \\
$h(\mathbf{x}^{new})$  & \multicolumn{1}{c|}{$\mathbf{x}^{old}$}   & CL-based adapter \cite{seo2025metric}                       & 63.14          & 50.01        & 89.10         & 78.20          & 80.50 & \multicolumn{1}{c|}{88.20}   &    77.58            & 72.14    \\
\rowcolor{gray!20} $h(\mathbf{x}^{new})$  & \multicolumn{1}{c|}{$\mathbf{x}^{old}$}   & \ours                               & 65.72               & 52.70               & 90.20                & 81.08               & 91.40       & \multicolumn{1}{c|}{92.30}       & 82.44                      & 75.36   \\
\midrule

\multicolumn{11}{c}{Hidden $\times 4$}                    \\ \midrule
$h(\mathbf{x}^{new})$  & \multicolumn{1}{c|}{$\mathbf{x}^{old}$}   & Base adapter \cite{FCT_side}                     & 56.32           & 50.52         & 85.20          & 79.46         & 78.60 & \multicolumn{1}{c|}{88.00}   & 73.37                & 72.66    \\
$h(\mathbf{x}^{new})$  & \multicolumn{1}{c|}{$\mathbf{x}^{old}$}   & EC-style adapter \cite{FCT_EC}                        & 58.42          & 50.33         & 86.00          & 79.62          & 78.70 & \multicolumn{1}{c|}{86.90}   & 74.37                & 72.28    \\
$h(\mathbf{x}^{new})$  & \multicolumn{1}{c|}{$\mathbf{x}^{old}$}   & CL-based adapter \cite{seo2025metric}                       & 64.00          & 49.98         & 88.90         & 78.26          & 81.20 & \multicolumn{1}{c|}{89.10}   & 78.03                & 72.45     \\
\rowcolor{gray!20} $h(\mathbf{x}^{new})$  & \multicolumn{1}{c|}{$\mathbf{x}^{old}$}   & \ours                        &       65.98    & 52.91      & 90.50          & 81.28          & 91.00 & \multicolumn{1}{c|}{92.00}   & 82.49               & 75.40     \\ \midrule

\multicolumn{11}{c}{Hidden $\times 8$}                    \\ \midrule
$h(\mathbf{x}^{new})$  & \multicolumn{1}{c|}{$\mathbf{x}^{old}$}   & Base adapter \cite{FCT_side}                     & 56.32           & 50.52         & 85.20          & 79.46         & 78.60 & \multicolumn{1}{c|}{88.00}   & 73.37                & 72.66    \\
$h(\mathbf{x}^{new})$  & \multicolumn{1}{c|}{$\mathbf{x}^{old}$}   & EC-style adapter \cite{FCT_EC}                        & 58.12           & 50.26         & 86.50          & 79.58          & 79.10 & \multicolumn{1}{c|}{87.10}   & 74.57               & 72.31   \\
$h(\mathbf{x}^{new})$  & \multicolumn{1}{c|}{$\mathbf{x}^{old}$}   & CL-based adapter \cite{seo2025metric}                       & 63.90          & 49.98         & 89.20         & 78.42         & 81.10 & \multicolumn{1}{c|}{87.60}   & 78.07                & 72.00     \\
\rowcolor{gray!20} $h(\mathbf{x}^{new})$  & \multicolumn{1}{c|}{$\mathbf{x}^{old}$}   & \ours                 &       66.42               & 52.94               & 90.70                & 81.24                & 91.40       & \multicolumn{1}{c|}{92.10}       & 82.84                     & 75.43    \\ 
\bottomrule

\end{tabular}
    
    }
    \label{tables/analyses_hidden}
\end{table}

\subsection{Different base adapters}

Since we employ a single adapter shared across all modalities, we compare this design against modality-specific adapters with various loss configurations in \cref{tables/analyses_3}. As shown, the unified single adapter achieves slightly better performance than separate adapters, leading us to adopt this design as our default. Additionally, $L2$ loss generally outperforms $L1$ loss. While the CL-based adapter performs competitively in the E5-V to UniME-7B setting, as observed in the main results, it excels primarily in T2I retrieval but falls short in I2T retrieval in other model variations. In contrast, our method consistently achieves superior performance across all scenarios.

\begin{table}[h!]
    \centering
     \caption{\textbf{Analyses on base adapters}. Details are the same as in \cref{tables/i2t_t2i_table}. }
     \resizebox{\columnwidth}{!}{
    \setlength{\tabcolsep}{3pt}
    
\begin{tabular}{ccc|cccccccc}
\midrule
\multicolumn{2}{c|}{\textbf{Case}}         & \multirow{2}{*}{\textbf{Variants}} & \multicolumn{2}{c}{\textbf{COCO}} & \multicolumn{2}{c}{\textbf{Flickr30K}} & \multicolumn{2}{c|}{\textbf{Urban1K}} & \multicolumn{2}{c}{\textbf{Average}} \\ \cline{4-11} 
query      & \multicolumn{1}{c|}{database} &                                  & I2T R@1         & T2I R@1         & I2T R@1          & T2I R@1          & I2T R@1 & \multicolumn{1}{c|}{T2I R@1} & I2T R@1                  & T2I R@1   \\ \midrule
\multicolumn{11}{c}{$\mathbf{x}^{old}$: E5-V-7B \cite{e5v}, $\mathbf{x}^{new}$: UniME-7B \cite{unime}}                                                                                                                                                                                        \\ \midrule
$h(\mathbf{x}^{new})$  & \multicolumn{1}{c|}{$\mathbf{x}^{old}$}   & Single L1-based adapter                     & 56.18               & 49.92               & 83.70               & 78.54                & 76.20       & \multicolumn{1}{c|}{87.20}       & 72.03                        & 71.89    \\
$h(\mathbf{x}^{new})$  & \multicolumn{1}{c|}{$\mathbf{x}^{old}$}   & Single L2-based adapter           & 55.90           & 50.21           & 85.1            & 79.22            & 77.80   & \multicolumn{1}{c|}{86.90}   & 72.93                    & 72.11      \\
$h(\mathbf{x}^{new})$  & \multicolumn{1}{c|}{$\mathbf{x}^{old}$}   & Single CL-based adapter                     & 63.18               & 50.12               & 88.70                & 78.46                & 81.70       & \multicolumn{1}{c|}{88.70}       & 77.86                        & 72.43       \\ \midrule

$h(\mathbf{x}^{new})$  & \multicolumn{1}{c|}{$\mathbf{x}^{old}$}   & Separate L1-based adapter                         & 55.42              & 49.72              & 83.60 & 78.34                & 76.50       & \multicolumn{1}{c|}{84.30}       & 71.84                       & 70.79    \\  
$h(\mathbf{x}^{new})$  & \multicolumn{1}{c|}{$\mathbf{x}^{old}$}  &  Separate L2-based adapter               & 55.20               & 49.72               & 84.20               & 78.78               & 77.50      & \multicolumn{1}{c|}{84.50}       & 72.30                        & 71.00    \\ 
$h(\mathbf{x}^{new})$  & \multicolumn{1}{c|}{$\mathbf{x}^{old}$}   &  Separate CL-based adapter            & 61.64               & 50.24               & 87.80                & 78.54               & 79.30       & \multicolumn{1}{c|}{88.70}       & 76.25                      & 72.49    \\ 
\bottomrule

\end{tabular}
    
    }
    \label{tables/analyses_3}
\end{table}
\subsection{Combination with CL-based adapter \cite{seo2025metric}} \label{sec:appendix_kd}
We additionally evaluate a variant in which a contrastive learning (CL) loss is used for the base adapter. Specifically, in place of the base adapter loss in \cref{eq:base_loss}, we use the CL objective defined in \cref{eq:cl_loss}. For the point-wise preservation loss, instead of the $L2$ formulation in \cref{eq:point_loss}, we adopt a CL-based variant that treats $(g(h(\mathbf{x}_i^{\text{new}})), \mathbf{x}_i^{\text{new}})$ as a positive pair. Note that the remaining preservation losses are applied in their original form.
As shown in \cref{tables/analyses_cl}, our method remains effective under this configuration, demonstrating the generalizability of our proposed loss design beyond the $L2$-based adapter.

\begin{table}[h!]
    \centering
     \caption{\textbf{Combination with CL-based adapter \cite{seo2025metric}}. Details are the same as in \cref{tables/i2t_t2i_table}. }
     \resizebox{\columnwidth}{!}{
    \setlength{\tabcolsep}{3pt}
    
\begin{tabular}{ccc|cccccccc}
\midrule
\multicolumn{2}{c|}{\textbf{Case}}         & \multirow{2}{*}{\textbf{Method}} & \multicolumn{2}{c}{\textbf{COCO}} & \multicolumn{2}{c}{\textbf{Flickr30K}} & \multicolumn{2}{c|}{\textbf{Urban1K}} & \multicolumn{2}{c}{\textbf{Average}} \\ \cline{4-11} 
query      & \multicolumn{1}{c|}{database} &                                  & I2T R@1         & T2I R@1         & I2T R@1          & T2I R@1          & I2T R@1 & \multicolumn{1}{c|}{T2I R@1} & I2T R@1                  & T2I R@1   \\ \midrule
\multicolumn{11}{c}{$\mathbf{x}^{old}$: CLIP ViT-L/14 \cite{clip}, $\mathbf{x}^{new}$: GME-7B \cite{gme}}                                                                                                                                                                                    \\ 
 \midrule
$h(\mathbf{x}^{new})$  & \multicolumn{1}{c|}{$\mathbf{x}^{old}$}   & CL-based adapter \cite{seo2025metric}                     & 54.38               & 37.06               & 76.00                & 68.00               & 63.40      & \multicolumn{1}{c|}{64.10}       & 64.59                        & 56.38      \\
\rowcolor{gray!20} $h(\mathbf{x}^{new})$  & \multicolumn{1}{c|}{$\mathbf{x}^{old}$}   & \ours (w/ CL)                        & 64.00         & 38.39         & 86.90          & 68.58         &71.40   & \multicolumn{1}{c|}{75.50}   & 74.10                 & 60.82  \\ \midrule

\multicolumn{11}{c}{$\mathbf{x}^{old}$: CLIP ViT-L/14 \cite{clip}, $\mathbf{x}^{new}$: UniME-7B \cite{unime}}                                                                                                       
\\ \midrule
$h(\mathbf{x}^{new})$  & \multicolumn{1}{c|}{$\mathbf{x}^{old}$}   & CL-based adapter \cite{seo2025metric}                     & 53.00               & 38.42               & 79.40               & 68.76               & 69.30       & \multicolumn{1}{c|}{70.70}       & 67.23                        & 59.29      \\
\rowcolor{gray!20} $h(\mathbf{x}^{new})$  & \multicolumn{1}{c|}{$\mathbf{x}^{old}$}   & \ours (w/ CL)                        & 65.70         & 40.73          & 91.00         & 70.54          & 79.90  & \multicolumn{1}{c|}{75.50}   & 78.87                    & 62.26    \\ \midrule

\multicolumn{11}{c}{$\mathbf{x}^{old}$: E5-V \cite{e5v}, $\mathbf{x}^{new}$: UniME-7B \cite{unime}}                                                                                                                                                                                        \\ \midrule

$h(\mathbf{x}^{new})$  & \multicolumn{1}{c|}{$\mathbf{x}^{old}$}   & CL-based adapter \cite{seo2025metric}                     & 63.18               & 50.12               & 88.70                & 78.46                & 81.70       & \multicolumn{1}{c|}{88.70}       & 77.86                        & 72.43      \\
\rowcolor{gray!20} $h(\mathbf{x}^{new})$  & \multicolumn{1}{c|}{$\mathbf{x}^{old}$}   & \ours (w/ CL)                        & 67.80         & 52.38         & 91.50          & 80.66         &91.00   & \multicolumn{1}{c|}{92.00}   & 83.43                   & 75.01   \\ 

\bottomrule

\end{tabular}
    
    }
    \label{tables/analyses_cl}
\end{table}

\subsection{Use of composition data} \label{sec:appendix_data}

As described in \cref{sec:experiments}, we additionally use composition data consisting of fused image–text inputs. Since dual-encoder models such as CLIP cannot directly process image–text compositions, we approximate them during training by summing the corresponding image and text features to construct a composite embedding.
As shown in \cref{tables/analyses_mmeb_it}, incorporating composition data improves the performance of $L2$- and EC-style adapters in the E5-V $\rightarrow$ UniME compatibility setting on the MMEB benchmark. We therefore adopt composition data as part of our default experimental setup to avoid underestimating baseline performance.
At the same time, as reported in \cref{tables/analyses_it}, our method consistently achieves considerable improvements both with and without composition data, indicating that its effectiveness does not rely on this design choice.

When using a dual-tower model (CLIP) as the old model in image–text benchmarks, we employ separate reverse adapters $g$ for each modality (image and text). The composition input is fed into the adapter for text modality.
Empirically, we observe that using separate modality-specific adapters $g$ versus a single shared reverse adapter $g$ results in only marginal performance differences. 
We use the modality-specific setup as our default, while noting that a single shared reverse adapter $g$ achieves comparable performance in most cases.
For the main adapter $h$, we always use a single unified adapter.

\begin{table}[h!]
    \centering
     \caption{\textbf{Inclusion of composition data.} Details are the same as in \cref{tables/mmeb_table}. }
     \resizebox{\columnwidth}{!}{
    \setlength{\tabcolsep}{3pt}
    \begin{tabular}{ccccccc|c}
\midrule
\multicolumn{2}{c|}{\textbf{Case}}         & \multicolumn{1}{c|}{\multirow{2}{*}{\textbf{Method}}} & \multicolumn{4}{c|}{\textbf{Per Meta-Task Score}}                & \multicolumn{1}{c}{\textbf{Average}}      \\ \cline{4-8} 
query      & \multicolumn{1}{c|}{database} & \multicolumn{1}{c|}{}                                 & Classification (10) & VQA (10) & Retrieval (12) & Grounding (4) & Overall (36) \\ \midrule

\multicolumn{8}{c}{w/o composition data}                                                                                                                                                        \\ \midrule

$h(\mathbf{x}^{new})$  & \multicolumn{1}{c|}{$\mathbf{x}^{old}$ }   & \multicolumn{1}{c|}{Base adapter \cite{FCT_side}}                     & 24.67               & 3.04    & 31.52         & 42.68       & 22.94        \\
$h(\mathbf{x}^{new})$  & \multicolumn{1}{c|}{$\mathbf{x}^{old}$ }   & \multicolumn{1}{c|}{EC-style adapter \cite{FCT_EC}}                     &  25.33             & 3.53    & 33.76       & 44.78        & 24.24      \\ \midrule

\multicolumn{8}{c}{w/ composition data}                                                                                                                                                        \\ \midrule

$h(\mathbf{x}^{new})$  & \multicolumn{1}{c|}{$\mathbf{x}^{old}$ }   & \multicolumn{1}{c|}{Base adapter \cite{FCT_side}}                     & 42.58               & 29.99    & 51.24          & 56.52         & 43.51       

\\
$h(\mathbf{x}^{new})$  & \multicolumn{1}{c|}{$\mathbf{x}^{old}$ }   & \multicolumn{1}{c|}{EC-style adapter \cite{FCT_EC}}                     & 43.23               & 30.91    & 51.72        & 56.83        & 44.14      \\

 \bottomrule

\end{tabular}
    
    }
    \label{tables/analyses_mmeb_it}
\end{table}
\begin{table}[h!]
    \centering
     \caption{\textbf{I2T/T2I results without composition data}. Details are the same as in \cref{tables/i2t_t2i_table}.}
     \resizebox{\columnwidth}{!}{
    \setlength{\tabcolsep}{3pt}
    \begin{tabular}{ccc|cccccccc}
\midrule
\multicolumn{2}{c|}{\textbf{Case}}         & \multirow{2}{*}{\textbf{Method}} & \multicolumn{2}{c}{\textbf{COCO}} & \multicolumn{2}{c}{\textbf{Flickr30K}} & \multicolumn{2}{c|}{\textbf{Urban1K}} & \multicolumn{2}{c}{\textbf{Average}} \\ \cline{4-11} 
query      & \multicolumn{1}{c|}{database} &                                  & I2T R@1         & T2I R@1         & I2T R@1          & T2I R@1          & I2T R@1 & \multicolumn{1}{c|}{T2I R@1} & I2T R@1                  & T2I R@1   \\ \midrule
\multicolumn{11}{c}{w/o composition data}                                                                                                                                                                                        \\ \midrule

$h(\mathbf{x}^{new})$  & \multicolumn{1}{c|}{$\mathbf{x}^{old}$}   & Base adapter \cite{FCT_side}                     & 55.90           & 50.21           & 85.10            & 79.22            & 77.80   & \multicolumn{1}{c|}{86.90}   & 72.93                    & 72.11     \\
$h(\mathbf{x}^{new})$  & \multicolumn{1}{c|}{$\mathbf{x}^{old}$}   & EC-style adapter \cite{FCT_EC}                      & 57.30               & 50.32               & 85.90               & 79.52                & 78.10       & \multicolumn{1}{c|}{88.10}       & 73.77                        & 72.65      \\
$h(\mathbf{x}^{new})$  & \multicolumn{1}{c|}{$\mathbf{x}^{old}$}   & CL-based adapter \cite{seo2025metric}                     & 63.78               & 50.21               & 88.50               & 78.68                & 82.90       & \multicolumn{1}{c|}{89.00}       & 78.39                     & 72.63      \\
\rowcolor{gray!20} $h(\mathbf{x}^{new})$  & \multicolumn{1}{c|}{$\mathbf{x}^{old}$}   & \ours                      & 65.38        & 52.78          & 90.40            & 80.96          & 90.30  & \multicolumn{1}{c|}{92.10}   & 82.03              & 75.28  \\  \midrule

\end{tabular}
    
    }
    \label{tables/analyses_it}
\end{table}

\subsection{Hyper-parameters} \label{sec:appendix_hyperparams}
\cref{tables/analyses_lambda} presents the full results corresponding to the figures in the main paper. As discussed therein, our method exhibits robust performance across varying values of $\lambda$, demonstrating its practical advantage for hyperparameter selection. We further evaluate performance with varying focal parameter $\gamma$ and different weightings among the preservation loss terms in \cref{tables/ablation_hyperparameter}. In the preservation loss weight ablation, $(a,b,c)$ denotes the weights assigned to the point-, distance-, and angle-wise preservation losses, respectively.
While some variation is observed, performance remains consistently strong across all configurations and continues to outperform the baselines by a clear margin.

\begin{table}[h!]
    \centering
     \caption{\textbf{Analyses on $\lambda$}. Details are the same as in \cref{tables/i2t_t2i_table}. }
     \resizebox{\columnwidth}{!}{
    \setlength{\tabcolsep}{3pt}
    
\begin{tabular}{ccc|cccccccc}
\midrule
\multicolumn{2}{c|}{\textbf{Case}}         & \multirow{2}{*}{\textbf{Variants}} & \multicolumn{2}{c}{\textbf{COCO}} & \multicolumn{2}{c}{\textbf{Flickr30K}} & \multicolumn{2}{c|}{\textbf{Urban1K}} & \multicolumn{2}{c}{\textbf{Average}} \\ \cline{4-11} 
query      & \multicolumn{1}{c|}{database} &                                  & I2T R@1         & T2I R@1         & I2T R@1          & T2I R@1          & I2T R@1 & \multicolumn{1}{c|}{T2I R@1} & I2T R@1                  & T2I R@1   \\ \midrule

\multicolumn{11}{c}{$\lambda$ ablation}                    \\ \midrule
$h(\mathbf{x}^{new})$  & \multicolumn{1}{c|}{$\mathbf{x}^{old}$}    & 0.0           & 55.90           & 50.21           & 85.10            & 79.22            & 77.80   & \multicolumn{1}{c|}{86.90}   & 72.93                    & 72.11      \\

$h(\mathbf{x}^{new})$  & \multicolumn{1}{c|}{$\mathbf{x}^{old}$}   & 0.1                    & 61.44           & 51.55         & 87.80          & 79.74          & 84.70 & \multicolumn{1}{c|}{90.10}   & 77.98                & 73.97     \\
$h(\mathbf{x}^{new})$  & \multicolumn{1}{c|}{$\mathbf{x}^{old}$}   & 0.5                       & 64.86          & 52.63        & 89.90         & 80.70         & 90.30 & \multicolumn{1}{c|}{92.00}   & 81.69                 & 75.11     \\
$h(\mathbf{x}^{new})$  & \multicolumn{1}{c|}{$\mathbf{x}^{old}$}   & 1.0                   & 65.28          & 52.61           & 90.60            & 81.10           & 90.20   & \multicolumn{1}{c|}{92.10}   & 82.03                    & 75.27    \\
$h(\mathbf{x}^{new})$  & \multicolumn{1}{c|}{$\mathbf{x}^{old}$}    & 2.0           & 65.42           & 52.80        & 90.00          & 81.24          & 90.60 & \multicolumn{1}{c|}{91.60}   & 82.01               & 75.21     \\
$h(\mathbf{x}^{new})$  & \multicolumn{1}{c|}{$\mathbf{x}^{old}$}    & 5.0           &  64.82         & 52.65         & 89.80          & 81.08          & 90.40 & \multicolumn{1}{c|}{91.70}   & 81.67                & 75.14     \\

$h(\mathbf{x}^{new})$  & \multicolumn{1}{c|}{$\mathbf{x}^{old}$}    & 10.0           & 64.64          & 52.46         & 90.10         & 81.02         & 90.40 & \multicolumn{1}{c|}{91.50}   & 81.71                & 74.99    \\
$h(\mathbf{x}^{new})$  & \multicolumn{1}{c|}{$\mathbf{x}^{old}$}    & 100.0           & 61.02           & 51.54         & 87.80          & 80.40          & 87.60 & \multicolumn{1}{c|}{90.10}   & 78.80                 & 74.01    \\

$h(\mathbf{x}^{new})$  & \multicolumn{1}{c|}{$\mathbf{x}^{old}$}    & 500.0           & 40.86           & 46.64         & 74.40        & 76.02         & 72.20 & \multicolumn{1}{c|}{80.10}   & 62.49                & 67.59     \\

$h(\mathbf{x}^{new})$  & \multicolumn{1}{c|}{$\mathbf{x}^{old}$}    & $\infty$           & 0.0          & 0.0         &  0.0            &  0.0            &  0.0   & \multicolumn{1}{c|}{ 0.0  }   & 0.0  &  0.0       \\

\bottomrule

\end{tabular}
    
    }
    \label{tables/analyses_lambda}
\end{table}

\begin{table}[h!]
    \centering
     \caption{\textbf{Additional hyper-parameter analyses.} $\gamma$ denotes the focal parameter defined in \cref{eq:point_focal_loss}. In the preservation-loss weight ablation, $(a,b,c)$ denotes the weights assigned to the point-, distance-, and angle-wise preservation losses, respectively. Other details are the same as in \cref{tables/i2t_t2i_table}.}
     \resizebox{\columnwidth}{!}{
    \setlength{\tabcolsep}{3pt}
    
\begin{tabular}{ccc|cccccccc}
\midrule
\multicolumn{2}{c|}{\textbf{Case}}         & \multirow{2}{*}{\textbf{Variants}} & \multicolumn{2}{c}{\textbf{COCO}} & \multicolumn{2}{c}{\textbf{Flickr30K}} & \multicolumn{2}{c|}{\textbf{Urban1K}} & \multicolumn{2}{c}{\textbf{Average}} \\ \cline{4-11} 
query      & \multicolumn{1}{c|}{database} &                                  & I2T R@1         & T2I R@1         & I2T R@1          & T2I R@1          & I2T R@1 & \multicolumn{1}{c|}{T2I R@1} & I2T R@1                  & T2I R@1   \\ \midrule

\rowcolor{gray!20}  $h(\mathbf{x}^{new})$  & \multicolumn{1}{c|}{$\mathbf{x}^{old}$}   & Original                     & 65.28          & 52.61           & 90.60            & 81.10           & 90.20   & \multicolumn{1}{c|}{92.10}   & 82.03                    & 75.27    \\ \midrule

\multicolumn{11}{c}{$\gamma$ ablation}                    \\ \midrule
$h(\mathbf{x}^{new})$  & \multicolumn{1}{c|}{$\mathbf{x}^{old}$}   & 0.5                    &  65.18        & 52.65        & 90.10          & 81.12          & 90.70 & \multicolumn{1}{c|}{92.00}   & 81.99                 & 75.26    \\
$h(\mathbf{x}^{new})$  & \multicolumn{1}{c|}{$\mathbf{x}^{old}$}    & 2.0           & 65.28           & 52.69      & 90.40          & 81.42          & 90.10 & \multicolumn{1}{c|}{91.70}   & 81.93                 & 75.27    \\
\midrule

\multicolumn{11}{c}{preservation loss weight ablation (a,b,c)}                    \\ \midrule

$h(\mathbf{x}^{new})$  & \multicolumn{1}{c|}{$\mathbf{x}^{old}$}   & (1.0, 5.0, 5.0)                     & 64.82           & 52.65        & 89.80          & 81.08        & 90.40  & \multicolumn{1}{c|}{91.70}   & 81.67            & 75.14     \\
$h(\mathbf{x}^{new})$  & \multicolumn{1}{c|}{$\mathbf{x}^{old}$}   & (5.0, 1.0, 5.0)                         & 64.74         & 52.72       & 89.90          & 81.10         & 90.60 & \multicolumn{1}{c|}{91.70}   & 81.75                & 75.17     \\
$h(\mathbf{x}^{new})$  & \multicolumn{1}{c|}{$\mathbf{x}^{old}$}    & (5.0, 5.0, 1.0)         & 64.84           & 52.61        & 90.80         & 81.22         & 90.20 & \multicolumn{1}{c|}{91.60}   & 81.95                & 75.14     \\

$h(\mathbf{x}^{new})$  & \multicolumn{1}{c|}{$\mathbf{x}^{old}$}   & (5.0, 1.0, 1.0)                     & 65.10           & 52.70        & 90.10          & 81.12          & 90.60 & \multicolumn{1}{c|}{92.00}   & 81.93                 & 75.27    \\
$h(\mathbf{x}^{new})$  & \multicolumn{1}{c|}{$\mathbf{x}^{old}$}   & (1.0, 5.0, 1.0)                         & 65.24         & 52.74       & 90.60          & 81.12          & 90.70 & \multicolumn{1}{c|}{92.20}   & 82.18                & 75.35     \\
$h(\mathbf{x}^{new})$  & \multicolumn{1}{c|}{$\mathbf{x}^{old}$}    & (1.0, 1.0, 5.0)         & 64.72           & 52.71        & 89.80         & 81.28          & 90.70 & \multicolumn{1}{c|}{91.40}   & 81.74                 & 75.13    \\

\bottomrule

\end{tabular}
    
    }
    \label{tables/ablation_hyperparameter}
\end{table}

\subsection{Exclusion of UniME in \cref{tables/doc_video}.} \label{sec:appendix_exclude}
UniME is excluded because, as a new model, it does not achieve better performance than potential old models (CLIP or E5-V) on the visual document and video benchmark; not a proper model update scenario.
We believe this is likely because UniME is built on a LLaVA-NeXT backbone, which is not explicitly pre-trained on video data, whereas GME leverages Qwen2-VL, which is pre-trained on both visual document and video data and further fine-tuned on larger-scale and more diverse datasets to serve as an embedding model.

\section{Comparison with XBT \cite{BCT_XBT}} \label{sec:appendix_xbt}

Before presenting our results, we describe XBT \cite{BCT_XBT} in more detail. XBT is a two-stage approach: (1) it first pre-trains an adapter using text-only data for training efficiency, and (2) it then LoRA-tunes the backbone of the new model using a contrastive learning objective.
In the first stage, XBT trains the adapter with the contrastive loss in \cref{eq:cl_loss}, where $(h(\mathbf{x}_i^{\text{new}}), \mathbf{x}_i^{\text{old}})$ is treated as a positive pair. XBT uses only text data for efficiency. After the adapter is trained, XBT proceeds to the second stage, where it LoRA-tunes the backbone of the new model while keeping the adapter frozen (except for the LayerNorm parameters). This LoRA-tuning uses a cross-modal contrastive learning objective that treats $(h(\mathbf{x}_{i,\mathrm{img}}^{\text{new}}),h(\mathbf{x}_{i,\mathrm{txt}}^{\text{new}}))$ as a positive pair.

We evaluate XBT in our MLLM setting using the LLaVA-LCS dataset for training and observe degraded performance compared to even base adapters. We additionally experiment with a larger dataset (CC3M+SBU, approximately 4M image–text pairs) to match the original training scale of XBT, using the cleaned captions for CC3M to more closely follow XBT’s default training setup. As shown in \cref{tables/analyses_XBT}, the performance is still even weaker than old embedding models. All reported results correspond to the best-performing configurations over sweeps of LoRA hyperparameters (including different ranks and $\alpha$ values) and learning rates.

\begin{table}[h!]
    \centering
     \caption{\textbf{Comparison with XBT.} Details are the same as in \cref{tables/i2t_t2i_table}. }
     \resizebox{\columnwidth}{!}{
    \setlength{\tabcolsep}{3pt}
    
\begin{tabular}{ccc|cccccc}
\midrule
\multicolumn{2}{c|}{\textbf{Case}}         & \multirow{2}{*}{\textbf{XBT train set}} & \multicolumn{2}{c}{\textbf{COCO}} & \multicolumn{2}{c}{\textbf{Flickr30k}} & \multicolumn{2}{c}{\textbf{Average}} \\ \cline{4-9} 
query      & \multicolumn{1}{c|}{database} &                                  & I2T R@1         & T2I R@1         & I2T R@1          & T2I R@1          & I2T R@1          & T2I R@1   \\ \midrule

$\mathbf{x}^{old}$     & \multicolumn{1}{c|}{$\mathbf{x}^{old}$}   & -                                & 62.24           & 50.98           & 88.20            & 80.64            & 75.22            & 65.81     \\
$\mathbf{x}^{new}$     & \multicolumn{1}{c|}{$\mathbf{x}^{new}$}   & -                                & 70.08           & 53.71           & 93.40            & 82.34            & 81.74            & 68.03     \\ \midrule
\multicolumn{9}{c}{XBT}                                                                                                                                                                 \\ \midrule
$h(\mathbf{x}^{new}_\text{lora})$  & \multicolumn{1}{c|}{$\mathbf{x}^{old}$}   & LLaVA-LCS                    & 60.06          & 48.62           & 86.30            & 76.26            & 73.18            & 62.44     \\
$h(\mathbf{x}^{new}_\text{lora})$  & \multicolumn{1}{c|}{$\mathbf{x}^{old}$}   & CC3M+SBU                      & 60.88           & 50.02           & 88.70            & 78.32           & 74.79           & 64.17     \\         
           \midrule
\multicolumn{9}{c}{Direct LoRA-tuning for new model backbone}                                                                                                                                                                 \\ \midrule
$\mathbf{x}^{new}_\text{lora}$  & \multicolumn{1}{c|}{$\mathbf{x}^{new}_\text{lora}$}   & LLaVA-LCS                    & 65.54          & 47.83          & 89.90           & 76.20            &  77.72           & 62.02   \\
$\mathbf{x}^{new}_\text{lora}$  & \multicolumn{1}{c|}{$\mathbf{x}^{new}_\text{lora}$}   & CC3M+SBU                      & 62.68          & 51.32           & 87.50            & 79.98           &  75.09         & 65.65   \\         
           \midrule

\end{tabular}
    
    }
    \label{tables/analyses_XBT}
\end{table}

We attribute this inferior performance of XBT in our MLLM setting to a fundamental difference in the difficulty of applying LoRA-tuning to CLIP- and MLLM-based embedding models. We hypothesize that the success of XBT in its original XBT paper setup (\ie, the CLIP-to-CLIP upgrade scenario) mainly arises from improving the new embedding model itself via LoRA-tuning. Specifically, as reported in \cite[Table 1]{BCT_XBT}, the retrieval performance of the LoRA-tuned new model (even when evaluated in the projected space) substantially exceeds that of the original new model, with gains of $+15.71$ in T2I R@1 and $+3.93$ in I2T R@1 on the NoCaps dataset.
Such improvements from LoRA-tuning with clean captions are plausible for CLIP, which was originally pre-trained on noisy, web-scale data. Prior work has shown that training with higher-quality captions can significantly improve performance \cite{blip,blip2}.

However, this strategy appears ineffective for MLLMs. MLLM-based embedding models are typically already obtained by LoRA tuning large base MLLMs on large-scale, carefully curated datasets using tailored training strategies such as hard-negative mining. Consequently, additional LoRA-tuning with a simple contrastive objective on moderate-scale datasets provides little benefit and can even be detrimental. 
To examine this, in \cref{tables/analyses_XBT} (rows 5 and 6), we directly apply LoRA fine-tuning to the new model backbone in its original embedding space, without using the adapter, which constitutes an easier setting for LoRA than LoRA-tuning in the projected space. As shown in \cref{tables/analyses_XBT} (rows 5 and 6), the resulting LoRA-tuned embeddings fail to improve performance and in some cases degrade it, highlighting the risk of further updating the backbone of MLLM-based embedding models.

Since LoRA fine-tuning degrades the performance of the new embedding model even in its own space, this degradation is likely to be amplified in the projected space used for backward compatibility. As a result, the projected embeddings become noisier inputs to the adapter, further compromising compatibility, consistent with our empirical findings. These results underscore the difficulty of applying additional LoRA tuning in MLLM settings: modifying the new model backbone risks degrading its original performance unless substantially more sophisticated data curation and training strategies are employed.
In contrast, our adapter-only approach achieves effective backward compatibility without requiring large-scale, carefully curated additional training data or risking degradation of the base model’s performance.

\section{Computational analyses}
In addition to adapter training, our pipeline requires a one-time preprocessing stage consisting of feature extraction and neighbor mining, which takes approximately 4–5 hours on 8 A100 GPUs.
 We argue that this is a highly efficient one-time investment when compared to the alternative of fine-tuning the backbone ({\it e.g.}, via LoRA). 
Once extracted, the MKP-Adapter can be trained very efficiently (approximately 1–2 hours). BCT approaches using LoRA instead require a forward and backward pass through the massive MLLM backbone for every training step. On the same 1M sample dataset, a single epoch of LoRA tuning exceeds 24 hours on 8 H100 GPUs.

As discussed in \cref{sec:analyses}, although our default setup uses 50 training epochs, our method achieves sufficient performance when trained for only 10 epochs, with a total training time of approximately 1–2 hours on a 1M-sample dataset in the CLIP ViT/L-14 $\rightarrow$ UniME (7B) setting. As shown in \cref{tables/10epoch_results}, while training for 50 epochs yields slightly better performance, the 10-epoch results are on par with those obtained with full 50-epoch training in most cases.

\begin{table}[h!]
    \centering
     \caption{\textbf{Comparison with 10 epoch results.} ``\ours (10E)'' denotes the results of \ours trained for 10 epochs. Other details are the same as in \cref{tables/i2t_t2i_table}.}
     \resizebox{\columnwidth}{!}{
    \setlength{\tabcolsep}{3pt}
    
\begin{tabular}{ccc|cccccccc}
\midrule
\multicolumn{2}{c|}{\textbf{Case}}         & \multirow{2}{*}{\textbf{Method}} & \multicolumn{2}{c}{\textbf{COCO}} & \multicolumn{2}{c}{\textbf{Flickr30K}} & \multicolumn{2}{c|}{\textbf{Urban1K}} & \multicolumn{2}{c}{\textbf{Average}} \\ \cline{4-11} 
query      & \multicolumn{1}{c|}{database} &                                  & I2T R@1         & T2I R@1         & I2T R@1          & T2I R@1          & I2T R@1 & \multicolumn{1}{c|}{T2I R@1} & I2T R@1                  & T2I R@1   \\ \midrule
\multicolumn{11}{c}{$\mathbf{x}^{old}$: CLIP ViT-L/14 \cite{clip}, $\mathbf{x}^{new}$: GME-7B \cite{gme}}                                                                                                                                                                                    \\ 
 \midrule
$h(\mathbf{x}^{new})$  & \multicolumn{1}{c|}{$\mathbf{x}^{old}$}   & \ours                   & 66.18          & 38.78          & 87.50           & 69.16            & 68.70   & \multicolumn{1}{c|}{77.60}   & 74.13                    & 61.85    \\
 $h(\mathbf{x}^{new})$  & \multicolumn{1}{c|}{$\mathbf{x}^{old}$}   & \ours (10E)                        & 65.64         & 38.40         & 87.20          & 68.26         &68.90   & \multicolumn{1}{c|}{74.60}   & 73.91                 & 60.42 \\ \midrule

\multicolumn{11}{c}{$\mathbf{x}^{old}$: CLIP ViT-L/14 \cite{clip}, $\mathbf{x}^{new}$: UniME-7B \cite{unime}}                                                                                                       
\\ \midrule
$h(\mathbf{x}^{new})$  & \multicolumn{1}{c|}{$\mathbf{x}^{old}$}   & \ours                        & 66.68          & 42.22           & 91.20           & 72.00           & 75.40   & \multicolumn{1}{c|}{75.00}   & 77.76                    & 63.07   \\
 $h(\mathbf{x}^{new})$  & \multicolumn{1}{c|}{$\mathbf{x}^{old}$}   & \ours (10E)                        & 66.10         & 42.30          & 91.00         & 71.18          & 74.80  & \multicolumn{1}{c|}{73.80}   & 77.30                    & 62.43    \\ \midrule

\multicolumn{11}{c}{$\mathbf{x}^{old}$: E5-V \cite{e5v}, $\mathbf{x}^{new}$: UniME-7B \cite{unime}}                                                                                                                                                                                        \\ \midrule

$h(\mathbf{x}^{new})$  & \multicolumn{1}{c|}{$\mathbf{x}^{old}$}   & \ours                       & 65.28          & 52.61           & 90.60            & 81.10           & 90.20   & \multicolumn{1}{c|}{92.10}   & 82.03                    & 75.27     \\
$h(\mathbf{x}^{new})$  & \multicolumn{1}{c|}{$\mathbf{x}^{old}$}   & \ours (10E)                         & 65.16         & 52.67         & 90.20          & 80.98         &90.70   & \multicolumn{1}{c|}{91.60}   & 82.01                   & 75.05   \\ 

\bottomrule

\end{tabular}
    
    }
    \label{tables/10epoch_results}
\end{table}

\subsection{Additional scaling experiments}
We conduct experiments by sub-sampling at varying scales (5k–500k) from LLaVA-LCS.
 As shown in \cref{tables/datascale}, the trend is consistent with \cref{tables/analyses_2}: a moderate dataset scale is sufficient, with performance saturating at around 200k samples.

\begin{table}[h!]
    \centering
     \caption{Additional data scale experiments with LLaVA-LCS. }
     \resizebox{0.8\columnwidth}{!}{
    {\large
    \begin{tabular}{c|cccccc}
\hline
E5-V$\rightarrow$UniME        & 5k & 10k & 50k & 100k & 200k & 500k       \\ \hline
I2T/T2I & 74.00/71.05     & 77.71/73.08      &  80.96/74.76    & 81.40/74.87     & 81.96/74.86 & 82.02/75.08 \\ \hline
\end{tabular}
    
    }
    }
    \label{tables/datascale}
\end{table}

\section{Additional evaluation} 
\subsection{Five random seed results} 
We additionally calculate results of ``MKP-Adapter'' and ``Base-adapter'' from five different random seeds under the same evaluation setup as \cref{tables/i2t_t2i_table}. The standard deviation of the average I2T/T2I R@1 is below $0.2\%$ points, showing that the observed improvement is consistent.

\subsection{Additional R@10 results corresponding to \cref{tables/i2t_t2i_table}} \label{subsec:appendix_R@10}
To provide a more comprehensive evaluation beyond R@1, in \cref{tables/i2t_t2i_R_10}, we report R@10 results under the same setting as \cref{tables/i2t_t2i_table}, where trends are consistent with R@1.

\begin{table}[h!]
    \centering
     \caption{\textbf{Additional R@10 results corresponding to \cref{tables/i2t_t2i_table}} }
     \resizebox{\columnwidth}{!}{
    \setlength{\tabcolsep}{3pt}
    
\begin{tabular}{ccc|cccccccc}
\midrule
\multicolumn{2}{c|}{\textbf{Case}}         & \multirow{2}{*}{\textbf{Method}} & \multicolumn{2}{c}{\textbf{COCO}} & \multicolumn{2}{c}{\textbf{Flickr30K}} & \multicolumn{2}{c|}{\textbf{Urban1K}} & \multicolumn{2}{c}{\textbf{Average}} \\ \cline{4-11} 
query      & \multicolumn{1}{c|}{database} &                                  & I2T R@10         & T2I R@10        & I2T R@10         & T2I R@10          & I2T R@10 & \multicolumn{1}{c|}{T2I R@10} & I2T R@10                  & T2I R@10   \\ \midrule
\multicolumn{11}{c}{$\mathbf{x}^{old}$: CLIP ViT-L/14 \cite{clip}, $\mathbf{x}^{new}$: GME-7B \cite{gme}}                                                                                                                                                                                    \\ 
 \midrule
$\mathbf{x}^{old}$     & \multicolumn{1}{c|}{$\mathbf{x}^{old}$}   & -                                & 86.46           & 71.07           & 99.20            & 92.16            & 94.00   & \multicolumn{1}{c|}{86.60}   & 93.22                  & 83.28     \\
$\mathbf{x}^{new}$     & \multicolumn{1}{c|}{$\mathbf{x}^{new}$}   & -                                & 93.14          & 87.38           & 99.80            & 97.44            & 99.40   & \multicolumn{1}{c|}{97.80}   & 97.45                     & 94.21     \\ \midrule
$h(\mathbf{x}^{new})$  & \multicolumn{1}{c|}{$\mathbf{x}^{old}$}   & Base adapter \cite{FCT_side}                     & 88.50           & 69.70           & 98.90             & 91.62            & 89.00   & \multicolumn{1}{c|}{82.60}   & 92.13                   & 81.31    \\
\rowcolor{gray!20} $h(\mathbf{x}^{new})$  & \multicolumn{1}{c|}{$\mathbf{x}^{old}$}   & \ours                         &  \textbf{91.80}          &  \textbf{72.90}          &  \textbf{99.20}           &  \textbf{93.62}            &  \textbf{92.70}   & \multicolumn{1}{c|}{ \textbf{95.50}}   &  \textbf{94.57}                    &  \textbf{87.34}    \\ \midrule

\multicolumn{11}{c}{$\mathbf{x}^{old}$: CLIP ViT-L/14 \cite{clip}, $\mathbf{x}^{new}$: UniME-7B \cite{unime}}                                                                                                       
\\ \midrule
$\mathbf{x}^{old}$     & \multicolumn{1}{c|}{$\mathbf{x}^{old}$}   & -                                & 86.46           & 71.07           & 99.20            & 92.16            & 94.00   & \multicolumn{1}{c|}{86.60}   & 93.22                   & 83.28    \\
$\mathbf{x}^{new}$     & \multicolumn{1}{c|}{$\mathbf{x}^{new}$}   & -                                 & 93.74           & 86.51           & 100.0            & 98.14            & 99.90   & \multicolumn{1}{c|}{99.50}   & 97.88                    & 94.72    \\ \midrule
$h(\mathbf{x}^{new})$  & \multicolumn{1}{c|}{$\mathbf{x}^{old}$}   & Base adapter \cite{FCT_side}                     & 84.86           & 73.09          & 98.60           & 93.74           & 86.00   & \multicolumn{1}{c|}{89.50}   & 89.82                   & 85.44    \\
\rowcolor{gray!20} $h(\mathbf{x}^{new})$  & \multicolumn{1}{c|}{$\mathbf{x}^{old}$}   & \ours                         & \textbf{92.50}         &  \textbf{77.16}          &  \textbf{99.60}           &  \textbf{95.20}           &  \textbf{94.70}   & \multicolumn{1}{c|}{ \textbf{95.00}}   &  \textbf{95.60 }                   &  \textbf{89.12}     \\ \midrule

\multicolumn{11}{c}{$\mathbf{x}^{old}$: E5-V \cite{e5v}, $\mathbf{x}^{new}$: UniME-7B \cite{unime}}                                                                                                                                                                                        \\ \midrule

$\mathbf{x}^{old}$     & \multicolumn{1}{c|}{$\mathbf{x}^{old}$}   & -                                & 90.52          & 83.89           & 99.60            & 97.64            & 98.40   & \multicolumn{1}{c|}{97.80}   & 96.17                    & 93.11  \\
$\mathbf{x}^{new}$     & \multicolumn{1}{c|}{$\mathbf{x}^{new}$}   & -                                 & 93.74           & 86.51           & 100.0            & 98.14            & 99.90   & \multicolumn{1}{c|}{99.50}   & 97.88                   & 94.72  \\ \midrule
$h(\mathbf{x}^{new})$  & \multicolumn{1}{c|}{$\mathbf{x}^{old}$}   & Base adapter \cite{FCT_side}                     & 87.42           & 83.28           & 99.20            & 97.32            & 97.50   & \multicolumn{1}{c|}{98.00}   & 94.71                & 92.87    \\
\rowcolor{gray!20} $h(\mathbf{x}^{new})$  & \multicolumn{1}{c|}{$\mathbf{x}^{old}$}   & \ours                      & \textbf{92.32}          & \textbf{85.25}           & \textbf{99.70}            & \textbf{97.78}           & \textbf{99.60}   & \multicolumn{1}{c|}{\textbf{99.20}}   & \textbf{97.21}                    & \textbf{94.08}    \\ 

\bottomrule

\end{tabular}
    
    }
    \label{tables/i2t_t2i_R_10}
\end{table}
\subsection{Large retrieval evaluation (COCO val entire set)}\label{sec:appendix_largecoco}
We also evaluate our method on a large-scale retrieval setting. In contrast to the standard MS-COCO Karpathy test split (5,000 images and 25,000 captions), we use the entire COCO 2014 validation set, comprising 40,373 images and 201,865 captions. Note that while the original validation set contains 40,504 images, we exclude images that do not have exactly five corresponding captions for a fair comparison. Our method again demonstrates strong performance in this large-scale setting, which more closely resembles real-world retrieval applications.

\begin{table}[h!]
    \centering
     \caption{\textbf{COCO val entire evaluation set.} }
     \resizebox{\columnwidth}{!}{
    \setlength{\tabcolsep}{3pt}
    
\begin{tabular}{ccc|cccccccc}
\midrule
\multicolumn{2}{c|}{\textbf{Case}}         & \multirow{2}{*}{\textbf{Method}} & \multicolumn{6}{c|}{\textbf{COCO val entire set}} & \multicolumn{2}{c}{\textbf{Average}} \\ \cline{4-11} 
query      & \multicolumn{1}{c|}{database} &                                  & I2T R@1         & I2T R@5      & I2T R@10      & T2I R@1      & T2I R@5 & \multicolumn{1}{c|}{T2I R@10} & I2T                  & T2I   \\ \midrule
\multicolumn{11}{c}{$\mathbf{x}^{old}$: CLIP-ViT/L14 \cite{clip}, $\mathbf{x}^{new}$: GME-7B \cite{gme}}                                                                                                                                                                                    \\ 
 \midrule
$\mathbf{x}^{old}$     & \multicolumn{1}{c|}{$\mathbf{x}^{old}$}   & -                                & 33.04           & 54.39           & 63.75            & 17.80           & 34.52   & \multicolumn{1}{c|}{42.91}   & 50.39                  & 31.74       \\
$\mathbf{x}^{new}$     & \multicolumn{1}{c|}{$\mathbf{x}^{new}$}   & -                              & 45.15         & 67.79           & 76.17         & 34.95          &  56.66   & \multicolumn{1}{c|}{65.43}   & 63.04                  & 52.35 \\ \midrule
$h(\mathbf{x}^{new})$  & \multicolumn{1}{c|}{$\mathbf{x}^{old}$}   & Base adapter \cite{FCT_side}                      & 34.12         & 57.02          & 66.30          & 17.66           &  33.97   & \multicolumn{1}{c|}{42.18}   & 52.48                  & 31.27   \\
$h(\mathbf{x}^{new})$  & \multicolumn{1}{c|}{$\mathbf{x}^{old}$}   & EC-style adapter \cite{FCT_EC}                 & 35.20         & 57.70          & 66.97          & 18.15           &  34.73   & \multicolumn{1}{c|}{43.12}   & 53.29                 & 32.00    \\
$h(\mathbf{x}^{new})$  & \multicolumn{1}{c|}{$\mathbf{x}^{old}$}   & CL-based adapter                     & 31.08          & 52.24           & 61.09         & 18.48          &  35.88   & \multicolumn{1}{c|}{44.76}   & 48.14                  & 33.04     \\
\rowcolor{gray!20} $h(\mathbf{x}^{new})$  & \multicolumn{1}{c|}{$\mathbf{x}^{old}$}   & \ours                         & \textbf{41.47}         & \textbf{64.42}          & \textbf{73.28}         & \textbf{19.68}         &  \textbf{37.08}   & \multicolumn{1}{c|}{\textbf{45.56}}   & \textbf{59.72}                  & \textbf{34.11}     \\ \midrule

\multicolumn{11}{c}{$\mathbf{x}^{old}$: CLIP-ViT/L14 \cite{clip}, $\mathbf{x}^{new}$: UniME-7B \cite{unime}}                                                                                                       
\\ \midrule
$\mathbf{x}^{old}$     & \multicolumn{1}{c|}{$\mathbf{x}^{old}$}   & -                                 & 33.04           & 54.39           & 63.75            & 17.80           & 34.52   & \multicolumn{1}{c|}{42.91}   &  50.39           & 31.74     \\
$\mathbf{x}^{new}$     & \multicolumn{1}{c|}{$\mathbf{x}^{new}$}   & -                                 & 46.37           & 69.11           & 77.44         &  31.32            & 53.52  & \multicolumn{1}{c|}{62.83}   & 64.31                  & 49.22    \\ \midrule
$h(\mathbf{x}^{new})$  & \multicolumn{1}{c|}{$\mathbf{x}^{old}$}   & Base adapter \cite{FCT_side}                     & 29.40           & 51.26        & 60.95         & 17.93          & 35.38  & \multicolumn{1}{c|}{44.34}   &   47.20              & 32.55    \\
$h(\mathbf{x}^{new})$  & \multicolumn{1}{c|}{$\mathbf{x}^{old}$}   & EC-style adapter \cite{FCT_EC}                 & 30.86       & 53.03        & 62.57          & 18.20          & 35.75  & \multicolumn{1}{c|}{44.62}   &   48.82           & 32.86    \\
$h(\mathbf{x}^{new})$  & \multicolumn{1}{c|}{$\mathbf{x}^{old}$}   & CL-based adapter                     & 32.88            & 54.91               & 64.01              & 16.99              & 34.53     & \multicolumn{1}{c|}{43.80}       & 50.60                 & 31.77 \\
\rowcolor{gray!20} $h(\mathbf{x}^{new})$  & \multicolumn{1}{c|}{$\mathbf{x}^{old}$}   & \ours                         & \textbf{41.49}          & \textbf{64.71}         & \textbf{73.92}         & \textbf{21.52}          &  \textbf{40.51}   & \multicolumn{1}{c|}{\textbf{49.74}}   & \textbf{60.04}               & \textbf{37.26}     \\ \midrule

\multicolumn{11}{c}{$\mathbf{x}^{old}$: E5-V \cite{e5v}, $\mathbf{x}^{new}$: UniME-7B \cite{unime}}                                                                                                                                                                                        \\ \midrule

$\mathbf{x}^{old}$     & \multicolumn{1}{c|}{$\mathbf{x}^{old}$}   & -                                & 37.11          &  60.04         & 69.26            & 28.84          & 50.18   & \multicolumn{1}{c|}{59.49}   &  55.47                  &  46.17    \\
$\mathbf{x}^{new}$     & \multicolumn{1}{c|}{$\mathbf{x}^{new}$}   & -                                 & 46.37           & 69.11           & 77.44         &  31.32            & 53.52  & \multicolumn{1}{c|}{62.83} &  64.31                  &  49.22 \\ \midrule
$h(\mathbf{x}^{new})$  & \multicolumn{1}{c|}{$\mathbf{x}^{old}$}   & Base adapter \cite{FCT_side}                     &  30.45           & 53.37           & 63.44            & 27.46           & 48.80   & \multicolumn{1}{c|}{58.28}   &   49.09             & 44.85    \\
$h(\mathbf{x}^{new})$  & \multicolumn{1}{c|}{$\mathbf{x}^{old}$}   & EC-style adapter \cite{FCT_EC}                      & 31.83               & 54.83               & 64.86              & 27.75                & 49.20       & \multicolumn{1}{c|}{58.67}       &  50.51                      & 45.21     \\
$h(\mathbf{x}^{new})$  & \multicolumn{1}{c|}{$\mathbf{x}^{old}$}   & CL-based adapter                     & 38.06             & 61.78               & 71.13                & 27.68               & 49.29      & \multicolumn{1}{c|}{58.79}       &  56.99               & 45.25   \\
\rowcolor{gray!20} $h(\mathbf{x}^{new})$  & \multicolumn{1}{c|}{$\mathbf{x}^{old}$}  & \ours                         & \textbf{40.61 }         &\textbf{ 64.25  }         & \textbf{ 73.38}         & \textbf{30.01}          &  \textbf{51.89}   & \multicolumn{1}{c|}{\textbf{61.30}}   & \textbf{59.41 }                 & \textbf{47.73}     \\  

\bottomrule

\end{tabular}
    
    }
    \label{tables/coco_val_whole}
\end{table}

\end{document}